\documentclass[runningheads]{article}
\usepackage[T1]{fontenc}
\usepackage{graphicx}
\usepackage{booktabs}
\usepackage[misc]{ifsym}

\usepackage{mwe}
\usepackage{verbatim} 
\usepackage{amsmath}
\usepackage{amsfonts}
\usepackage{amssymb}
\usepackage[colorlinks=true, allcolors=blue]{hyperref}
\usepackage{array}
\usepackage{enumitem}
\usepackage[english]{babel}
\usepackage[caption=false]{subfig}
\usepackage{placeins}
\usepackage{multirow}
\usepackage[margin=1in]{geometry}

\newcommand{\R}{\mathbb{R}}
\newcommand{\set}[1]{\mathcal{#1}}
\newcommand{\vect}[1]{\boldsymbol{#1}}
\newcommand{\graph}[1]{\mathcal{#1}}
\newcommand{\M}[2]{#2^{#1}}

\begin{document}

\title{Semantic Depth Matters: Explaining Errors of Deep Vision Networks through Perceived Class Similarities}

\date{}
\author{
  Katarzyna Filus\thanks{\texttt{kfilus@iitis.pl}},\quad 
  Michał Romaszewski,\quad 
  Mateusz Żarski\\[0.5em]
  Institute of Theoretical and Applied Informatics,\\
  Polish Academy of Sciences,\\
  Bałtycka 5, 44-100 Gliwice, Poland
}

\maketitle              
\begin{abstract}

Understanding deep neural network (DNN) behavior requires more than evaluating classification accuracy alone; analyzing errors and their predictability is equally crucial. Current evaluation methodologies lack transparency, particularly in explaining the underlying causes of network misclassifications. To address this, we introduce a novel framework that investigates the relationship between the semantic hierarchy depth perceived by a network and its real-data misclassification patterns. Central to our framework is the Similarity Depth (SD) metric, which quantifies the semantic hierarchy depth perceived by a network along with a method of evaluation of how closely the network's errors align with its internally perceived similarity structure. We also propose a graph-based visualization of model semantic relationships and misperceptions. A key advantage of our approach is that leveraging class templates -- representations derived from classifier layer weights -- is applicable to already trained networks without requiring additional data or experiments. 
Our approach reveals that deep vision networks encode specific semantic hierarchies and that high semantic depth improves the compliance between perceived class similarities and actual errors.

\end{abstract}

\section{Introduction}
Deep neural networks have revolutionized countless fields in computer vision. However, a significant challenge still lies in understanding the rationale for their decision-making. At the same time, this understanding is crucial for reliability and transparency given the growing responsibility delegated to the intelligent systems. While much effort is put to create more accurate systems and explain their single decisions \cite{bib:yuan2021explaining}, more emphasis should be placed on network errors. It is because it is not the correct predictions that make the network's decisions intuitive, but precisely their mistakes.

There is increasing demand for methods with built-in interpretability alongside accuracy metrics. Cognitive psychology offers relevant concepts, such as similarities and hierarchies, which are related to categorization \cite{bib:rosch1978cognition}. Although initially explored in network evaluation, research shifted primarily towards accuracy-driven performance improvements \cite{bib:russakovsky2015imagenet,bib:bertinetto2020making}. We argue these concepts merit renewed attention, as simple metrics (accuracy, loss) provide limited insights into how networks perceive class similarities and their impact on decisions and errors -- a shortcoming highlighted by recent studies \cite{bib:huang2021semantic,bib:bilal2017convolutional,bib:bertinetto2020making}. While existing work acknowledges that models frequently confuse similar classes \cite{bib:mopuri2020adversarial,bib:filus2024similarity}, the degree to which perceived similarity affects errors and their predictability remains underexplored.

Motivated by these deficiencies, we propose a novel framework for deep vision networks that inspects how deep, in the sense of semantic hierarchy, the perception of deep vision networks is. We propose to use the Similarity Depth (SD) metric that measures the depth of a network's semantic perception in comparison to a known class hierarchy (e.g., via WordNet). By connecting the model's perceived similarity with WordNet's semantic similarity, SD can be used as a descriptor of the network's semantic structure with an emphasis on the vertical dimension in the concept hierarchy. We also introduce graph-based compliance metrics for comparing class similarity structure obtained from confusion matrices or directly from model weights. When this compliance is high, model errors align with its perceived class similarities. We also propose an intuitive visualization tool that enables verification of the numerical results obtained from the framework and shows how it can be used to gain a deeper understanding of the perceived model class structure. The proposed framework makes classification results interpretable and provides clear insights into how the network's perception aligns with semantic categorizations.
Through experiments on well-known vision neural networks, we show that their perception of semantic similarity, obtained based on visual data only, is specific far beyond general concepts. We also show that compliance of model errors with their highest perceived similarities is proportional to the network semantic depth.

Key contributions of this work composing the proposed explainability framework are:
\begin{itemize}[noitemsep]
    \item We propose \textit{Similarity Depth (SD)}, a measure of network semantic depth defined relative to a known concept hierarchy. Using this measure, we experimentally demonstrate that the perceptual representation of established vision neural networks extends beyond general-level concepts. The proposed method requires no input data and can be applied directly to pretrained models relying solely on the analysis of network weights.
    \item We introduce \emph{Similarity Graph Compliance}, a metric quantifying the alignment between model errors and perceived class similarities derived directly from network weights. Our experiments indicate that similarity graph compliance is proportional to the network's semantic depth, enabling the analysis of model errors independently of input data.
    \item We present a visualization method that illustrates the similarity structure and hierarchical relationships within the model-perceived class space. We demonstrate how this visualization aids in explaining model decisions and identifying model deficiencies.
\end{itemize}

Our framework can help in `debugging' the errors of visual classifiers and provide deeper insights into both training datasets and model behavior. We will provide a repository with our implementation upon acceptance.

\section{Related Work}
Similarity and semantic hierarchies are vital tools in cognitive sciences to examine human concept representations~\cite{bib:goldstone1994similarity,bib:rosch1978cognition}. Similarity is considered in vertical and horizontal dimensions~\cite{bib:rosch1978cognition}: vertical captures categorical hierarchy, whereas horizontal reflects pairwise category relationships. The vertical dimension thus effectively assesses semantic understanding depth in deep networks~\cite{bib:bertinetto2020making}.
Semantic similarity, often relating to stable lexical structures like WordNet~\cite{bib:miller1990introduction}, serves as a standard reference when comparing with neural representations \cite{bib:filus2024similarity,bib:bilal2017convolutional,bib:bertinetto2020making}. However, alternatives like human ratings are also employed \cite{bib:muttenthaler2024improving}. 
Network-derived similarities typically use stimulus-based methods involving extracted features or confusion matrices \cite{bib:huang2021semantic,bib:kornblith2019similarity,bib:kriegeskorte2008rdm,bib:williamsequivalence}. Recently, image-free methods leveraging classifier weights have gained traction due to computational efficiency, particularly for adversarial evaluation metrics \cite{bib:filus2023netsat,bib:mopuri2020adversarial}.
Representing inter-class similarities as affinities within similarity graphs provides structured insight into concept relationships. Graph-based methods, widely used for analyzing internal neural connectivity and performance prediction \cite{bib:you2020graph}, are less common in examining high-level concepts for explainability, with few exceptions utilizing stimulus-based representations \cite{bib:purvine2023experimental}. Clustering and community detection effectively identify category relationships within  graphs \cite{bib:lawson2012population}, but their use in explainability remains limited. Existing metrics mainly emphasize horizontal similarities (pairwise relations) \cite{bib:mopuri2020adversarial,bib:filus2023netsat,bib:bilal2017convolutional}, with fewer addressing hierarchical depth through common ancestors for assessing dataset difficulty \cite{bib:deng2010does} or severity of mistakes \cite{bib:bertinetto2020making}. Visual examination commonly targets image-level region activations \cite{bib:zintgraf2017visualizing,bib:selvaraju2016grad}, with limited studies addressing whole-category representations \cite{bib:bilal2017convolutional}.

\begin{figure}[h!]
    \centering
    \includegraphics[width=0.9\textwidth]{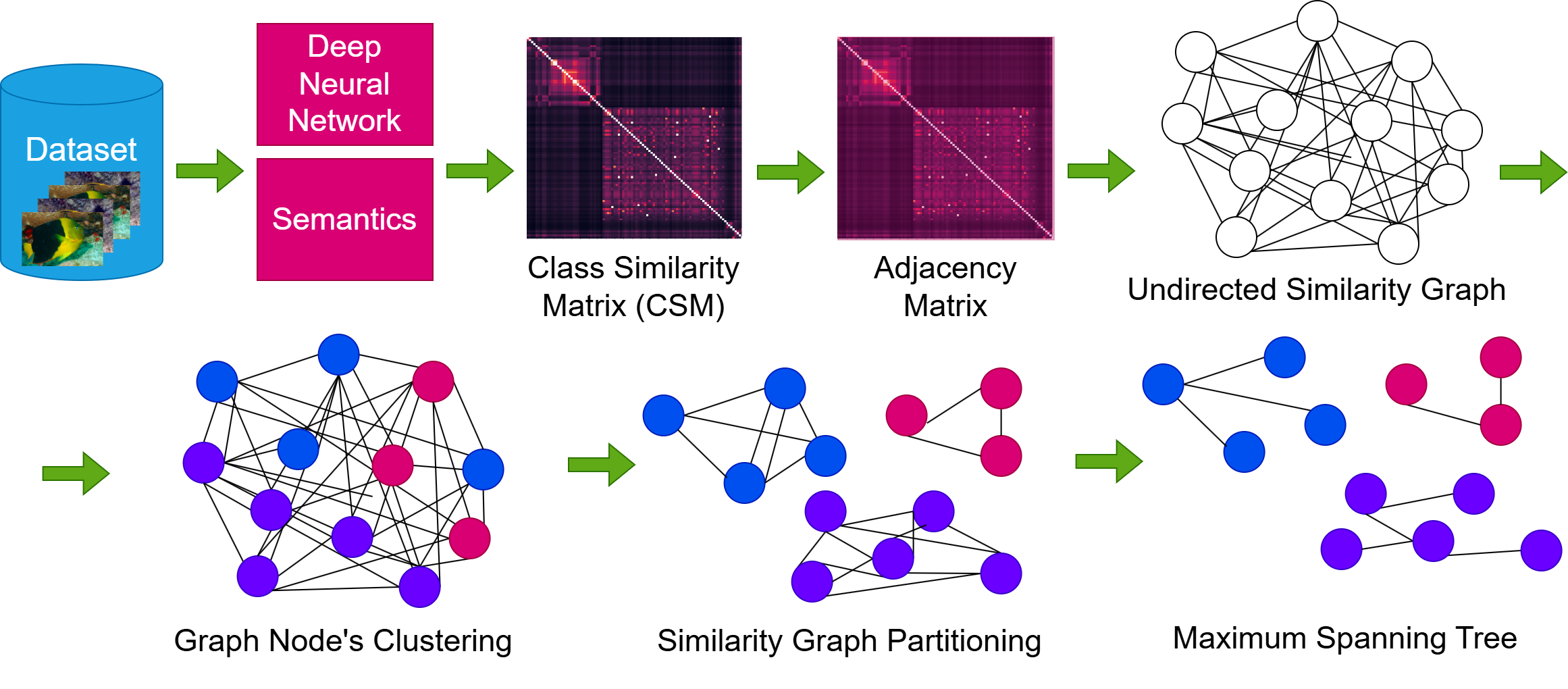}
    \caption{Overview of the main pipeline of the proposed framework. Class Similarity Matrices are transformed into undirected similarity graphs. The graph is then partitioned to obtain subgraphs with strong inter-node relations. Maximum Spanning Tree is then determined for all the subgraphs. These structures can be used to compute our Semantic Depth or for visualization explanations.  }
    \label{fig:pipeline}
\end{figure}

\section{Method}
\subsection{Semantic Perception Framework}

Our proposed framework addresses three aspects of network-perceived semantics: (1) estimation of the network's semantic depth perception (the depth of the perceived class hierarchy), (2) evaluation of the predictability of network errors (the alignment between the most frequent classification errors and perceived class similarities) and (3) visual assessment of correspondence between maximum perceived similarities and semantic similarities. By integrating these three aspects, our framework uniquely connects different notions of similarity, employing WordNet semantic similarity as an intuitive reference to clarify certain network-perceived similarities.

\begin{figure}[h!]
    \centering
    \subfloat[Similarity Depth metric]{%
        \includegraphics[width=0.4\textwidth]{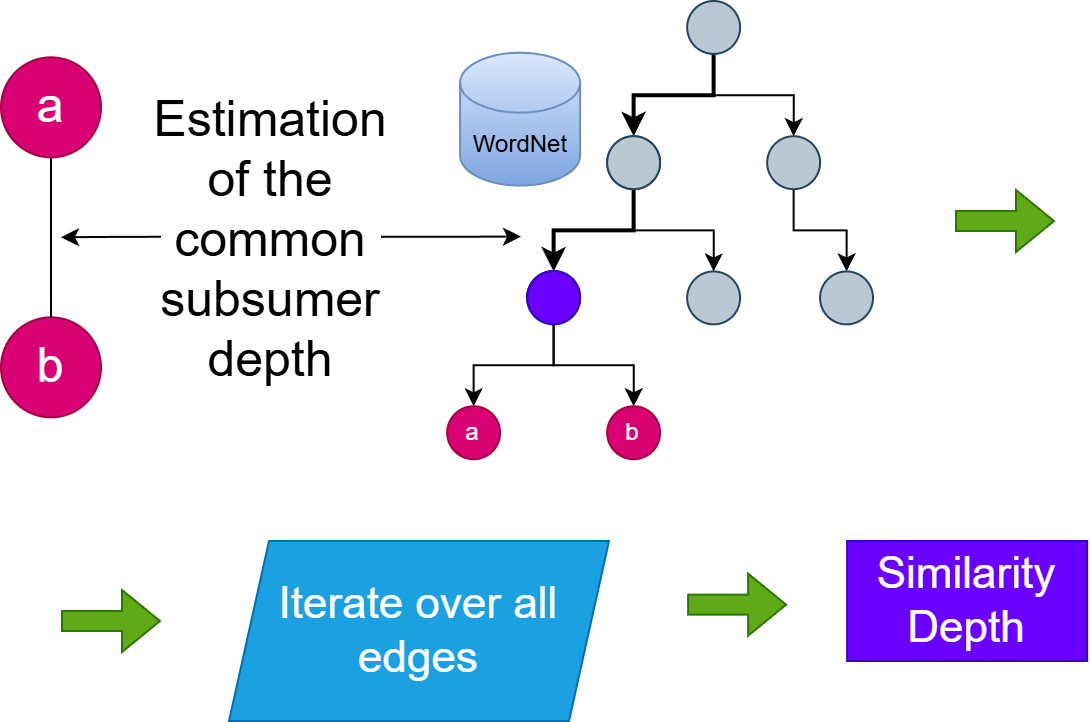}\label{subfig:sd}
    }
    \quad
    \subfloat[Explanation of graph edges]{%
        \includegraphics[width=0.4\textwidth]{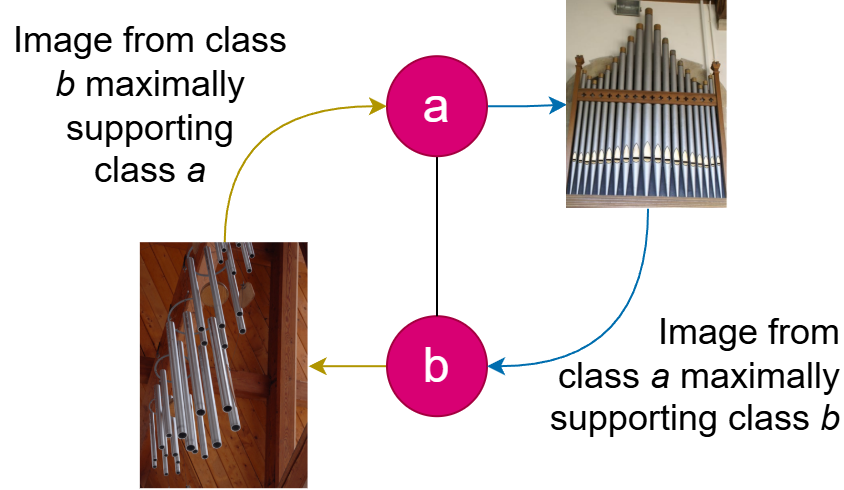}\label{subfig:xai}
    }
    \caption{Overview of the proposed quantitative and qualitative methods.}
    \label{fig:metric_xai}
\end{figure}

\begin{figure}[h!]
    \centering
    \includegraphics[width=0.9\textwidth]{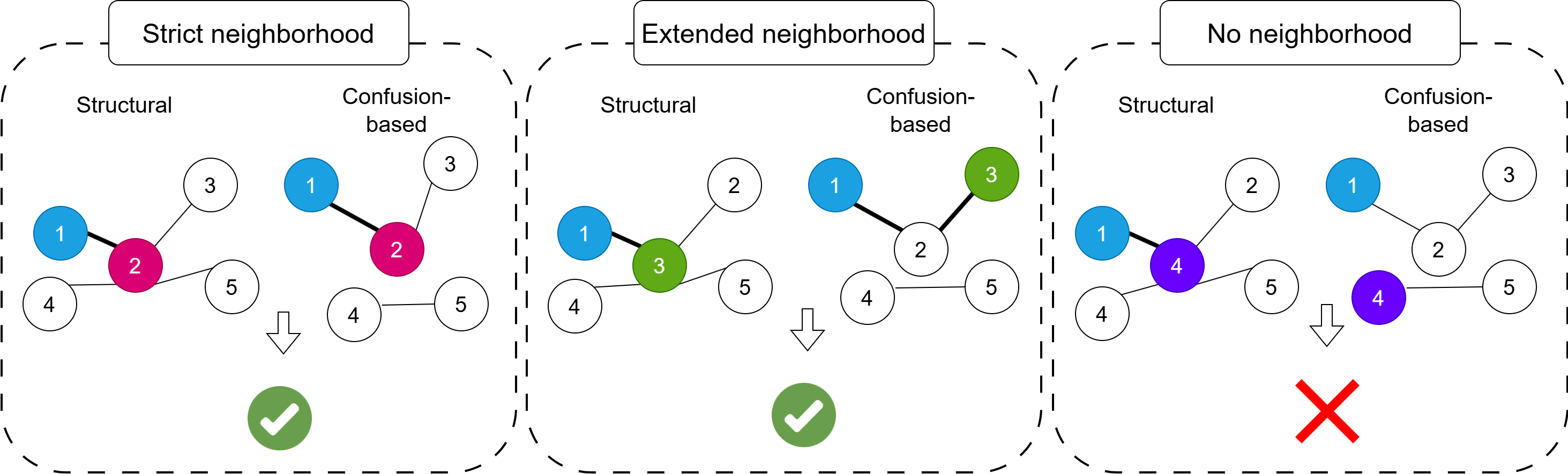}
    \caption{Different neighborhood types used in the graphs' compliance metrics.  }
    \label{fig:neighborhood}
\end{figure}


\subsection{Class Similarity Graphs}
\label{section:csg}
Class Similarity Graphs (CSG) provide a representation of relationships between the known classes. Given a large set of concepts $\set{Z}$, a set of classes $c\in\set{C}\subset\set{Z}$ can be represented by nodes of undirected, weighted graphs $\graph{G}=\langle\set{C},\set{E},\vect{S}\rangle$, where the non-negative, symmetric Class Similarity Matrices (CSM) $\vect{S}\in\M{\set{C}\times\set{C}}{\R_+}$ unambiguously define sets of edges storing all pairwise similarities between classes $\set{E}=\{e:e\in\set{C}\times\set{C}\land\vect{S}(e)>0\}$, where $\vect{S}(e)$ denotes weight of a weight matrix $\vect{S}$ for an edge $e$ (note that this matrix is symmetric). Initially, we assume that these graphs are fully connected, but they can be sparsified as zero values in CSM implicate that the corresponding edge does not exist.

We define three types of CSG:
\begin{itemize}
    \item \textbf{Semantic Similarity Graph (SSG)}: based on semantic similarity computed using the WordNet \cite{bib:miller1990introduction} ontology-based distances (we use Path length \cite{bib:pedersen2004wordnet}, Wu-Palmer - WUP \cite{bib:wu:1994verb} and Leacock-Chodorow \cite{bib:leacock1998combining} - LCH). 
    \item \textbf{Network Structural Similarity Graph (NSSG)}: Derived without input data, from the weights of the neural network. To obtain weight matrices, we use a final classification layer of a deep network, which weights can be treated as learned class representations \cite{bib:mopuri2020adversarial}. To do this, each neuron of the classification layer is assumed to be corresponding to one of the classes $c\in\set{C}$. A vector of weights $\vect{w}_c\in\R^{|\set{C}|}$ connecting  $c$-th neuron to the penultimate layer is treated as a representation of $c$ in the last layer's feature space. Similarity between these `templates' is computed with cosine similarity. Computing the similarities between all class pairs and scaling them to range $\langle 0, 1 \rangle$ results in the Structural \textbf{CSM}.
    \item \textbf{Network Functional Similarity Graph (NFSG)}: Computed from model outputs, where similarities reflect the functional behavior of the network (and indirect similarity). Let $ \vect{M} $ be the model confusion matrix, where $ \vect{M}_{ij} $ is the number of instances of class $ c_i $ that are classified by the model as class $ c_j $. To create a class similarity matrix $\vect{S}$ from matrix $ \vect{M} $, we first normalize each row of $ \vect{M} $ (so that rows sum to $1$), and then fill the diagonal with value $1$. This results in a functional class similarity matrix $\vect{S}$. Note that as this matrix could be asymmetric, meaning that $ \vect{S}_{ij} \neq \vect{S}_{ji} $, we ensure its symmetry by averaging all elements $ \vect{S}_{ij}$ and $\vect{S}_{ji} $. 
\end{itemize}

To refine the graph structure and focus on the strongest connections, we apply a clustering algorithm (Louvain, Affinity, OPTICS, HDBSCAN clustering) to partition the graph $\graph{G}$ into communities (connected components). Within each partition, we extract a Maximum Spanning Tree (MST) using Kruskal’s algorithm. The final CSG exhibits the following properties: every class remains connected within its partition (connected component); the MST guarantees that each pair of classes is connected through a unique path (no cycles); each connected component represents community corresponding to a meta-concept common to many classes.  By reducing the complexity and path redundancy, CSGs provide an efficient and intuitive way to analyze and understand neural network class similarities. We present these concepts graphically in Fig. \ref{fig:pipeline}.

\subsection{Similarity Depth (SD)}
WordNet organizes nouns in a hypernym–hyponym hierarchy (ISA hierarchy) ranging from very general concepts (e.g., `entity') at the top to highly specific ones (e.g., Irish wolfhound') at the bottom~\cite{bib:miller1990introduction}. It is generally assumed that synsets residing deeper in the WordNet hierarchy are semantically closer~\cite{bib:wang2011refining}. Consequently, depth within the ISA hierarchy serves as an indicator of both semantic similarity and concept specificity, underpinning various similarity metrics such as Wu and Palmer similarity~\cite{bib:wu:1994verb}.

Given a pair of synsets $(c_1,c_2)$, we define their shared depth as 
\begin{equation}
d(c_1,c_2)=\text{depth}(\text{LCS}(c_1,c_2)),
\end{equation}
where `$\text{depth}$' represents synset depth in the ISA hierarchy, and `LCS' denotes the least common subsumer, defined as the most specific shared ancestor for a pair of synsets. For a similarity graph $\graph{G}=\langle\set{C},\set{E},\vect{S}\rangle$, where vertices $\set{C}$ represent synsets and edges $\set{E}$ represent similarities between them, the average depth of the graph is calculated by averaging the depth values of synset pairs connected by edges. We present this concept graphically in Fig. \ref{subfig:sd}.

\subsection{Visual Similarity Graph Explanation}
Once the subgraphs are constructed and the depths of all LCSs are computed, this information is utilized to generate a structured and interpretable visualization. The edges of the graph are first represented using a heatmap, where the color hue corresponds to normalized similarity within the current graph. Each edge is further annotated with the depth of the common ancestor of the connected vertices in WordNet, previously used to compute the \textit{Semantic Depth (SD)} metric. Edges with a more general common ancestor (closer to the root, indicating lower depth) correspond to lower semantic similarity between classes. Such cases may reveal potential labeling inconsistencies or the influence of alternative semantic relationships—such as shared environments or reliance on purely visual fine-grained features rather than semantic attributes.

To further analyze these divergences, we then extract a representative subset of images that characterize the most extreme relationships. Given the graph $\graph{G}=\langle\set{C},\set{E},\vect{S}\rangle$ as defined in Section~\ref{section:csg}, an edge $e\in\set{E}$ can be defined by a pair of indices $(i, j)$ in the matrix $\vect{S}$. These indices correspond to classes in the set $\set{C}$. We extract training images from the corresponding classes $c_i$ and $c_j$. Let $\set{X}_i = \{ x_{i,1}, x_{i,2}, ..., x_{i,n} \}$ and $\set{X}_j = \{ x_{j,1}, x_{j,2}, ..., x_{j,m} \}$ be the sets of training images for $i$ and $j$, respectively.

Using the network's feature extractor $\phi(\cdot)$, we compute the feature representations of all the chosen images:
\begin{equation}
    \vect{z}_{i,k} = \phi(x_{i,k}), \quad \vect{z}_{j,l} = \phi(x_{j,l}), \quad \forall x_{i,k} \in X_i, \quad x_{j,l} \in X_j.
\end{equation}

Next, we use the previously defined weights of the final classifier as class templates as the mean feature representations, $\vect{w}_i$ and $\vect{w}_j$ for classes $c_i$ and $c_j$ respectively. To determine the most and least supporting images (images from class $c_i$ supporting class $c_j$ and vice versa), we compute the similarity between each instance and the weight template of the opposite class using a cosine similarity, and for each image $x_{i,k}$ obtain similarity $s_{i,k}^{(j)}$ from class $c_i$, and for each image $l$ from class $c_j$ - $s_{j,l}^{(i)}$. We then select:
\begin{itemize}
    \item Two most supporting images for a given edge (i, j): $x_{i,k^*}$ and $x_{j,l^*}$, where
    \begin{equation}
        k^* = \arg\max_{k} s_{i,k}^{(j)}, \quad l^* = \arg\max_{l} s_{j,l}^{(i)}.
    \end{equation}
    \item Two least supporting images: $x_{i,k^\dagger}$ and $x_{j,l^\dagger}$, where $k^\dagger$ and $l^\dagger$ are computed in the same way as $k^*$ and $l^*$ but with $\arg\min$ instead of $\arg\max$.
\end{itemize}

By analyzing these 4 images, we can identify underlying relationships causing the observed high similarity and explain them (e.g. common environments or shared visual traits). We present the graphical representation of the analysis of the supporting images in Fig. \ref{subfig:xai}.

\subsection{Measuring the Compliance of Structure-based and Confusion-based Graphs}

\textbf{Similarity Graph Compliance (SGC)} is a metric that measures the alignment between two similarity graphs, in particular, between similarity graphs of a given network created based on its internal representations' similarity or similarity estimated based on its classification errors. High alignment between these two similarity graphs means that the network's explicit internal similarity representation (direct estimation of similarity) matches its implicit judgments revealed by its mistakes (indirect estimation), which mirrors human cognitive consistency \cite{bib:goldstone1994similarity}. Since similarity representation can be derived directly from weights, if these two estimates align well, the network's errors can be inferred from its internal structure without experiments and data. This allows the evaluation of an aligned network in an ad-hoc manner directly from a trained model. High alignment also suggests that the network's internal class representations capture the key features of the underlying task performance and error patterns. In contrast, low alignment indicates that the model's internal representations do not adequately reflect the reasons for its mistakes.

SGC quantifies the degree of overlap between the edges of two graphs: the Network Structural Similarity Graph (NSSG) and the Network Functional Similarity Graph (NFSG). Following the description in the Section~\ref{section:csg}, given a source graph $\graph{G}_s=\langle\set{C},\set{E}_s,\vect{S}_s\rangle$ and a target graph $\graph{C}_t=\langle\set{V},\set{E}_t,\vect{S}_t\rangle$, the two compliance variants are defined as follows:
\begin{itemize}
    \item{\textbf{Strict Neighborhood Compliance}} evaluates the fraction of edges in the source graph that are also present in the target graph as is defined:
\begin{equation}
    \text{SGC}_{\text{strict}}(\graph{G}_s, \graph{G}_t) = \frac{|\set{E}_s \cap \set{E}_t|}{|\set{E}_s|}.
\end{equation}

\item{\textbf{Extended Neighborhood Compliance}}
allows edges from the source graph to be considered as present in the target graph if there exists a connecting path between the same vertices in $\graph{G}_t$, and is suitable for graphs with smaller subgraphs. Given an edge $e\in\set{E}_s$ between a pair of vertices $(v_i,v_j)\in\set{C}\times\set{C}$, 
its value is defined:
\begin{equation}
    \text{SGC}_{\text{extended}}(\graph{G}_s, \graph{G}_t) = \frac{| \{e: e \in \set{E}_s \wedge \exists \text{ path from } v_i \text{ to } v_j \text{ in } \graph{G}_t\} |}{|\set{E}_s|}.
\end{equation}
\end{itemize}

Note that SGC has obvious properties: its value is in the range $\langle0,1\rangle$,\\ $\text{SGC}_{\text{strict}}(\graph{G}_s, \graph{G}_t)=1$ implies both graphs have similar sets of edges and\\ $\text{SGC}_{\text{extended}}(\graph{G}_s, \graph{G}_t)=1$  implies both graphs have similar connected components.

The direction of comparison (choice of the source and target graph) implies a different interpretation of  results:
\begin{itemize}
    \item \textbf{source: NSSG (structural), target: NFSG (functional)} -  ratio of perceived similarities leading to confusion (\textbf{similarities causing confusions}).
    \item \textbf{source: NFSG (functional), target: NSSG (structural)} ratio of confusions that can be explained with perceived similarity (\textbf{confusions explained by similarities}).
\end{itemize}
Higher SGC values indicate greater structural similarity between the two graphs' edge sets and imply the emergence of similar meta-classes (class-based communities). Structural similarity leads to higher predictability of the network's decision-making processes through its mistakes and, thus, a more intuitive behavior. SGC can be analyzed independently or in relation to \textit{Semantic Depth (SD)} to study how semantic depth influences predictability and network behavior. See Fig. \ref{fig:neighborhood} for a graphical representation of different types of neighborhoods.

\subsection{Experimental Procedure}

\begin{table}[h]
\centering
\caption{Models considered in this study}\label{tab:model_list}
\begin{tabular}{ll}
\hline
\textbf{Model Type} & \textbf{Architectures} \\
\hline
Convolutional Neural Networks (CNNs) & Mobilenet V2~\cite{mobilenet_model}, ResNet 18~\cite{resnet_model},  \\
 & DenseNet 121~\cite{densenet_model}, EfficientNet~\cite{efficientnet_model},  \\
 & ConvNeXt~\cite{convnext_model} \\
Vision Transformers (ViTs) & ViT B~\cite{vit_model}, SWIN T~\cite{swin_model} \\
Hybrid model & MaxVit~\cite{maxvit_model} \\
\hline
\end{tabular}
\end{table}

Our training procedure (excluding the learning rate for transformer models) remained consistent to ensure reproducibility. We trained the models for 300 epochs with a batch size of 128. For optimization, we used AdamW optimizer~\cite{adamW} and added additional learning rate chained schedulers. For the warmup period, we used a linear scheduler for 10 epochs, and then \textit{Reduce on Plateau}~\cite{rop_scheduler} scheduler with a 20-epoch patience period and a reduce factor of 0.8. 

All our experiments were performed on a Linux-based system with Nvidia A100 GPGPU. 

In our experiments with different training paradigms than the supervised one, we use DINOv2 \cite{bib:oquab2023dinov2} trained with a Self-Supervised procedure and an image-text model - CLIP \cite{bib:radford2021learning}. We take the pre-trained models from: DINOv2: \url{https://huggingface.co/docs/transformers/model_doc/dinov2} and CLIP: \url{https://huggingface.co/docs/transformers/model_doc/clip}. \newline

\textbf{Datasets used}

The main body of our experiments was performed on two commonly used image classification datasets -- Mini-ImageNet~\cite{mini_imagenet} and CIFAR-100~\cite{cifar}. Mini-ImageNet dataset consists of 50000 training images and 10000 testing images, evenly distributed across 100 classes. It includes groups of objects in categories such as animals (e.g., tiger, whale), man-made objects (e.g. dome, tank), plants, and food entities. CIFAR-100 consists of 60000 32x32 px color images divided evenly into 100 classes, with entity labels similar to Mini-ImageNet. There are 600 images per class, with 500 images for training and 100 for testing.

In our experiments, we also performed a number of augmentations during the training process. In order to ensure a consistent shape of the input we resized the images to 224x224 px and applied a series of transformations for both training and testing images. For training, we used random horizontal flip and random rotation along with Gaussian blur and Gaussian noise and applied random perspective and random affine augmentations. For testing images we used inference resize augmentation with center crop and data normalization. \newline

\textbf{Other tools used}

For computing semantic similarities with the following distance measures: path length (path) \cite{bib:pedersen2004wordnet}, Wu and Palmer (WUP) \cite{bib:wu:1994verb}, Leacock and Chodorow (LCH) \cite{bib:leacock1998combining}, and thus obtaining the Class Similarity Matrices, we use the \textbf{NLTK} framework~\cite{bird2009natural} along with WordNet linguistic taxonomy. We compute semantic similarity for each pair of classes in the used datasets.

Although Mini-ImageNet labels are pre-encoded into Wordnet-based IDs, CIFAR-100 labels are not. Thus we prepared CIFAR-100 dataset labels in the same way they are encoded in Mini-ImageNet, having in mind the correct meaning of each label. In some cases, where the default WordNet meaning did not coincide with the actual object in the dataset images (such as plate (dish) encoded by default as home-plate (baseball base)), we corrected the label encoding manually to match the object in an image.

\section{Results}

In this section, we present the results obtained on eight state-of-the-art models trained on Mini-ImageNet. 
\paragraph{Semantic depth of tested models} Average SD values obtained for the tested networks are presented in Table~\ref{tab:combined_SD}.
As expected, SD computed directly from WordNet similarity measures (first three rows) consistently achieved high SD values (approximately 8-9), indicating highly specific semantic relations. 

Neural networks demonstrated lower semantic depths (approximately 7-8). To contextualize these values, Table~\ref{table:example_classes_mini_image_net} provides illustrative examples from mini-ImageNet. Semantic depth values 7-8 convey detailed and specific semantic information, whereas shallow depths indicate meaningless or excessively general similarities. This aligns closely with the observations reported in~\cite{bib:wang2011refining}. Its authors noted that for shallow LCS depths (less than 5), concept similarity is independent of depth. However, for depths beyond 5, common ancestors correlate with increased concept similarity. This supports our findings that neural networks tested in this study, on average, operate at a semantic depth indicative of specific semantic similarities likely correlating with human perception. 

\begin{table}
	\centering
	\caption{Semantic Depth (SD) for different clustering approaches.}
	\label{tab:combined_SD}
	\scriptsize
	\begin{tabular}{l|l|l|l|l|l|l}
		\textbf{Dataset} & \textbf{Network} & \textbf{Louvain} & \textbf{OPTICS} & \textbf{Affinity} & \textbf{HDBSCAN} & \textbf{No clustering} \\
		\hline
		\multirow{11}{*}{mini-ImageNet} & WordNet Path & 8.96 (15) & 8.48 (2) & 8.90 (20) & 8.39 (4) & 8.52 (1) \\
		& WordNet WUP & 8.65 (3) & 8.57 (5) & 9.05 (14) & 8.56 (4) & 8.56 (1) \\
		& WordNet LCH & 8.59 (3) & 8.42 (5) & 8.90 (14) & 8.39 (4) & 8.52 (1) \\
		& DenseNet121 & 7.36 (10) & 7.13 (1) & 6.66 (14) & 7.20 (3) & 7.13 (1) \\
		& EfficientNetV2 & 7.61 (6) & 7.56 (2) & 7.68 (16) & 7.34 (3) & 7.62 (1) \\
		& MobileNetV2 & 7.32 (6) & 7.34 (1) & 7.47 (15) & 7.36 (3) & 7.34 (1) \\
		& ConvNeXt & 7.51 (8) & 7.33 (1) & 7.29 (15) & 7.33 (1) & 7.33 (1) \\
		& ResNet152 & 7.57 (9) & 7.29 (1) & 7.20 (15) & 7.27 (3) & 7.29 (1) \\
		& MaxVitT & 7.43 (16) & 7.08 (1) & 7.12 (17) & 6.68 (3)  & 7.08 (1) \\
		& Swin-base & 7.65 (7) & 7.58 (1) & 7.66 (18) & 7.39 (4) & 7.58 (1) \\
		& ViTB & 7.70 (8) & 7.64 (1) & 7.41 (18) & 7.44 (3) & 7.64 (1) \\
	\end{tabular}
\end{table}

\paragraph{The impact of clustering} We observed that clustering generally increased the average semantic depth compared to graphs with a single connected component. In particular, the highest value of average depth always resulted from clustering. We observed that clustering sometimes mitigates meaningless connections arising from the greedy nature of Kruskal’s algorithm used for spanning tree generation. Our observations suggest that Affinity Propagation, due to its interpretability and positive impact on SD values, is a promising method of choice for generating smaller and more manageable subgraphs.

\begin{figure}[h!]
    \centering
    \subfloat[200 - strict]{%
        \includegraphics[width=0.45\textwidth]{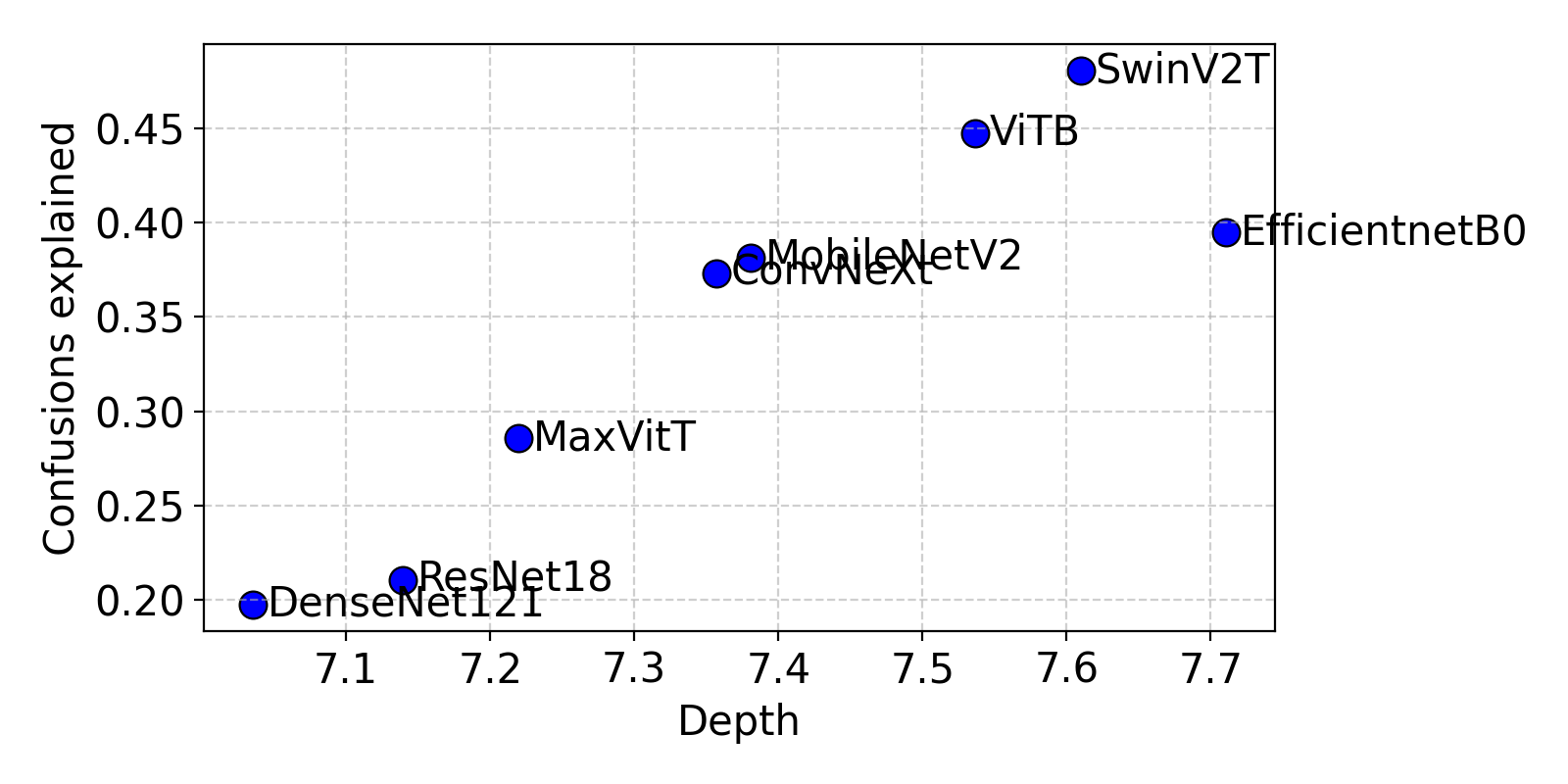}
    }
    \subfloat[299 - strict]{%
        \includegraphics[width=0.45\textwidth]{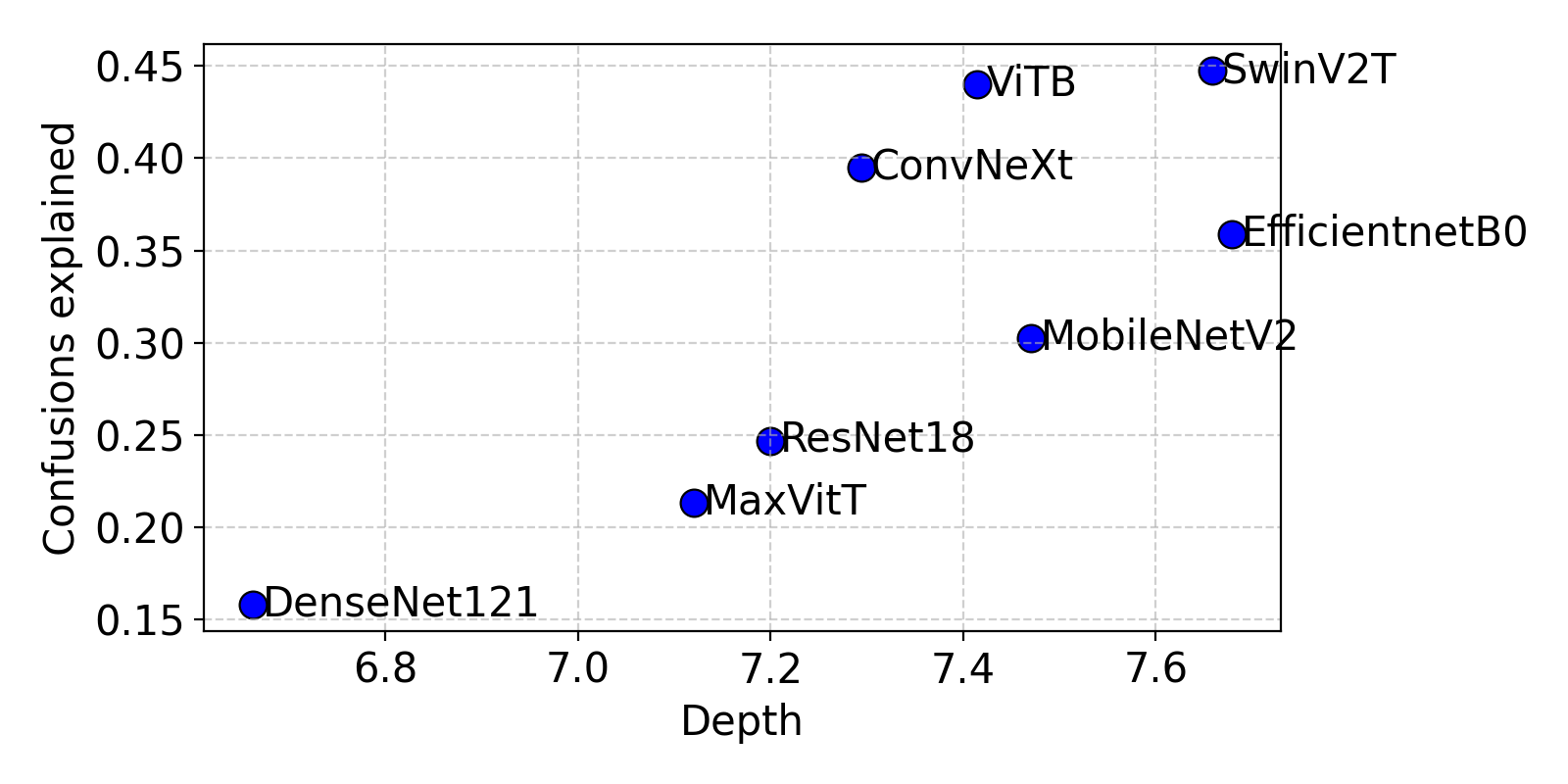}
    }
\\[-3ex]
    \subfloat[200 - extended]{%
        \includegraphics[width=0.45\textwidth]{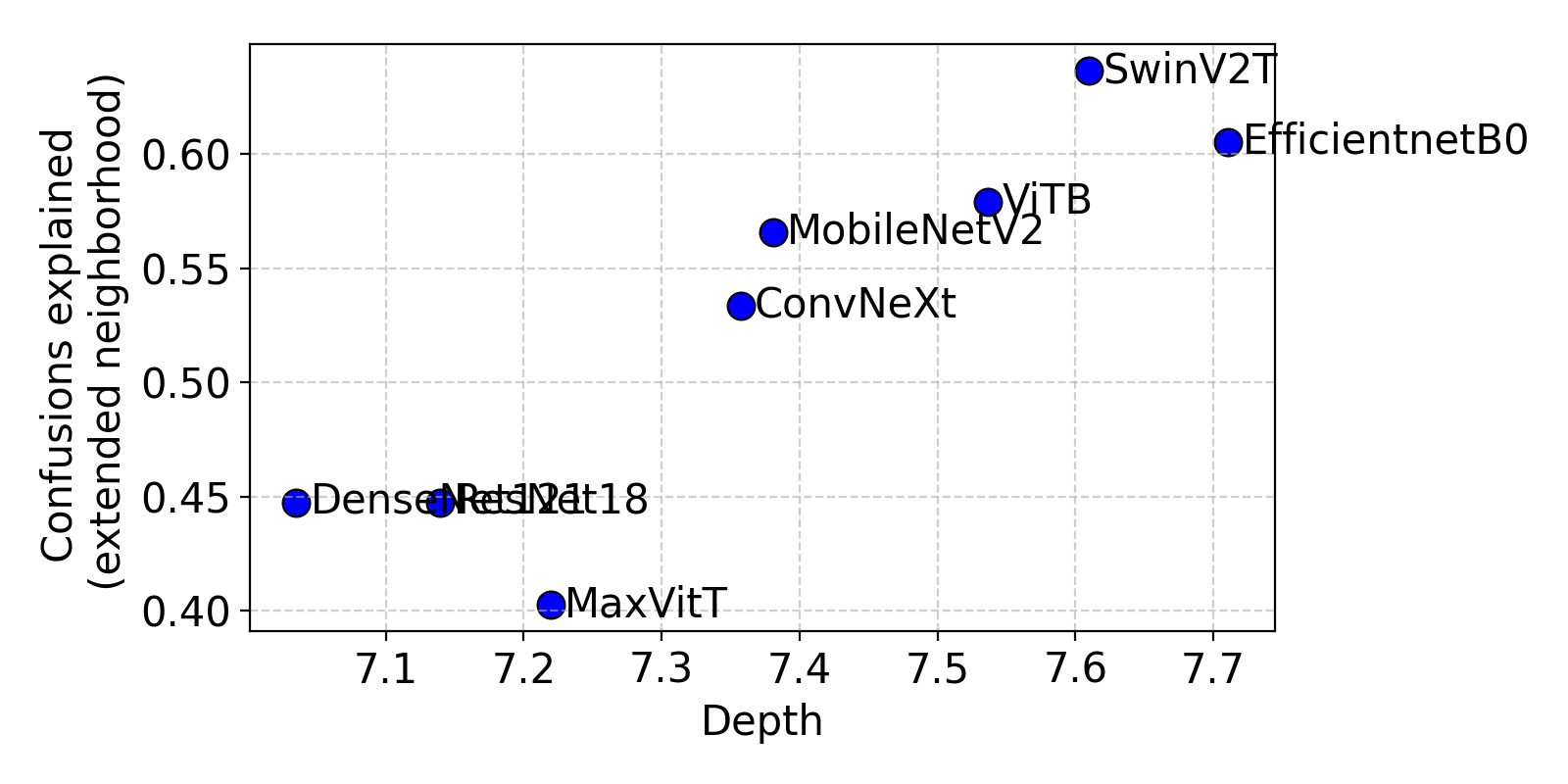}
    }
    \subfloat[299 - extended]{%
        \includegraphics[width=0.45\textwidth]{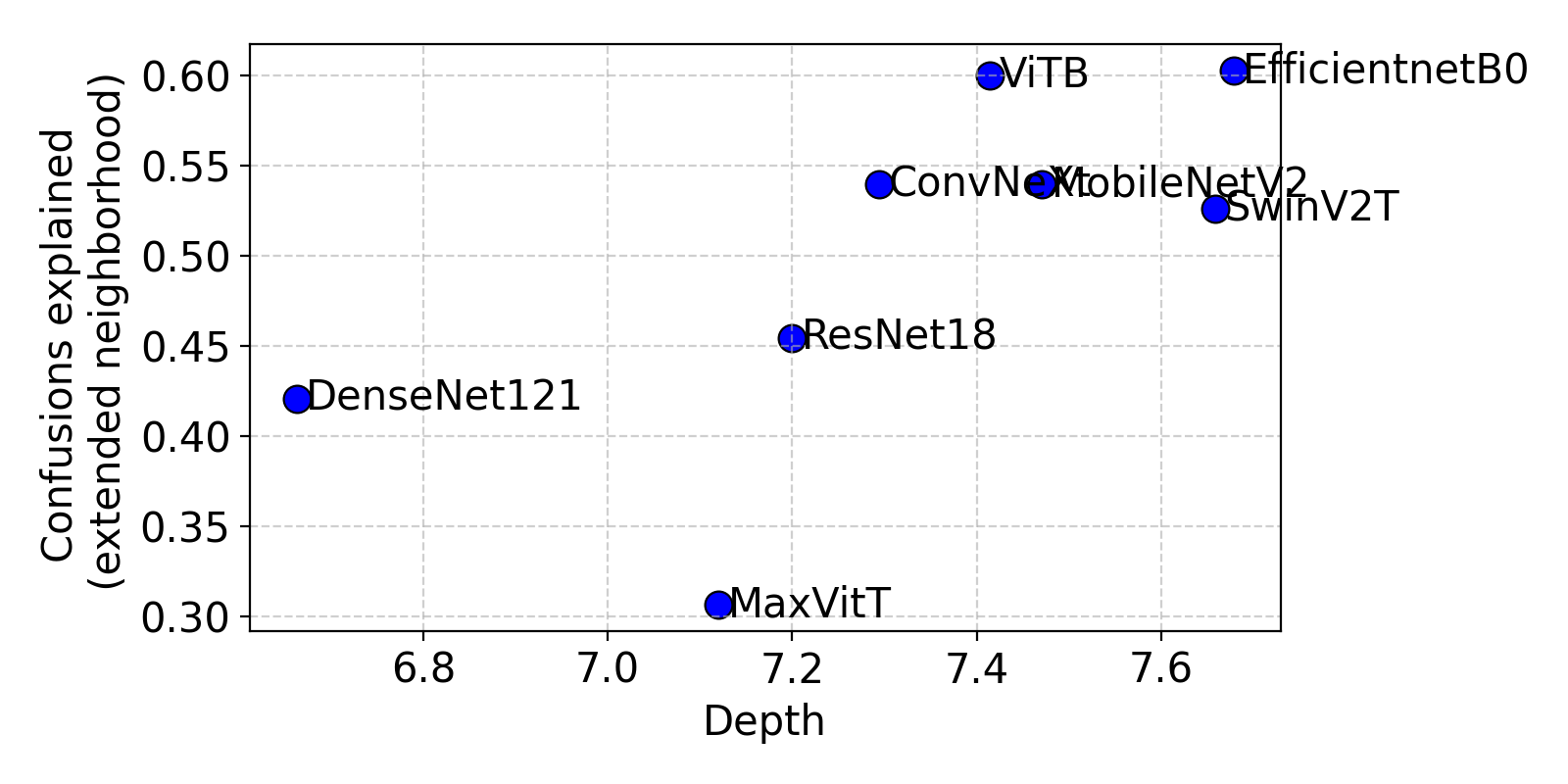}
    }
    \caption{Similarity Graph Compliance (\textbf{SGC}) as a function of \textbf{SD}: source - NFSG (functional), target - NSSG (structural). Mistakes we can explain with  similarities. }
    \label{fig:mini_confusions}
\end{figure}

\begin{figure}[h!]
    \centering
    \subfloat[200 - strict]{%
        \includegraphics[width=0.45\textwidth]{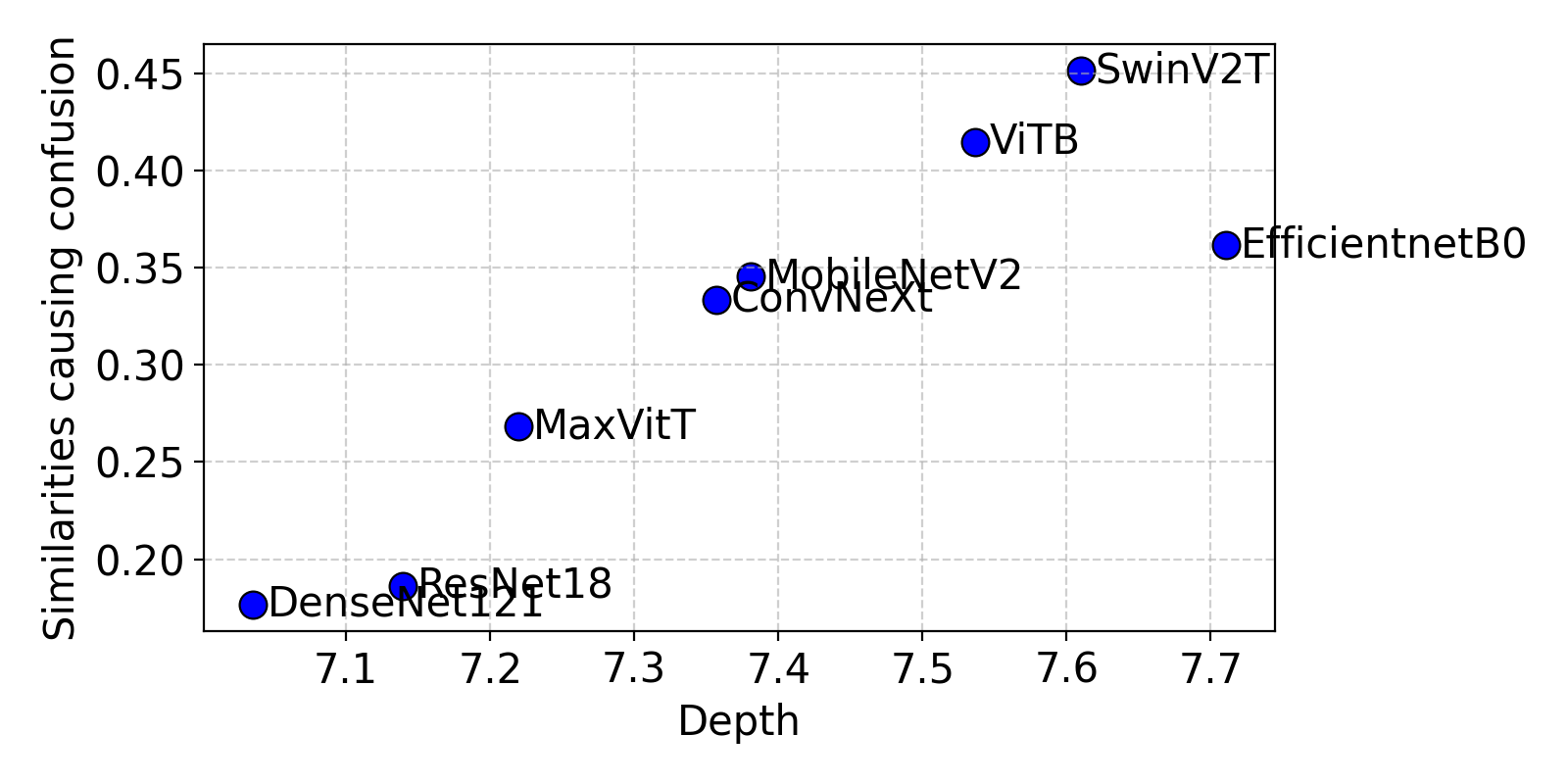}
    }
    \subfloat[299 - strict]{%
        \includegraphics[width=0.45\textwidth]{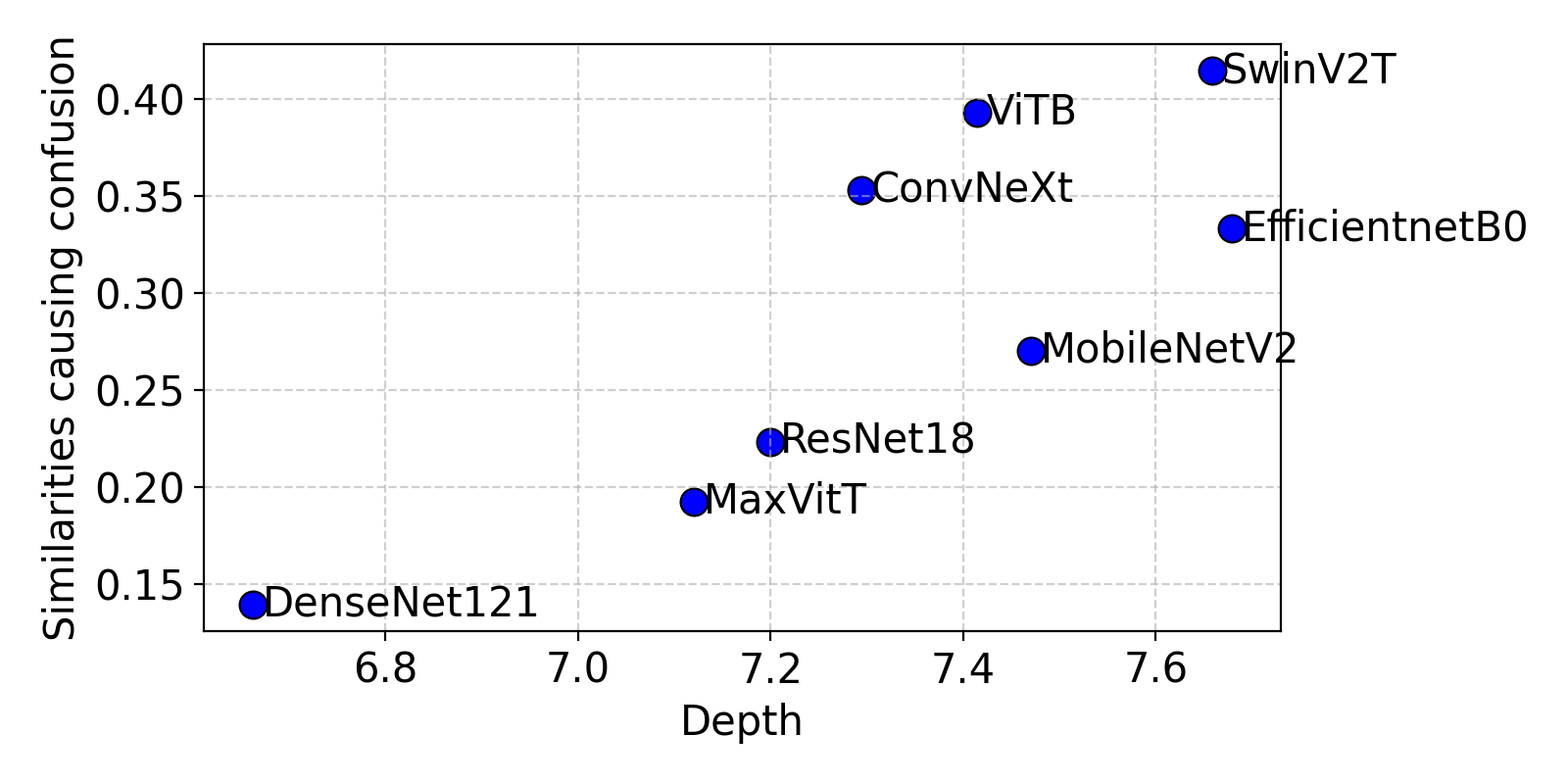}
    }
\\[-3ex]
    \subfloat[200 - extended]{%
        \includegraphics[width=0.45\textwidth]{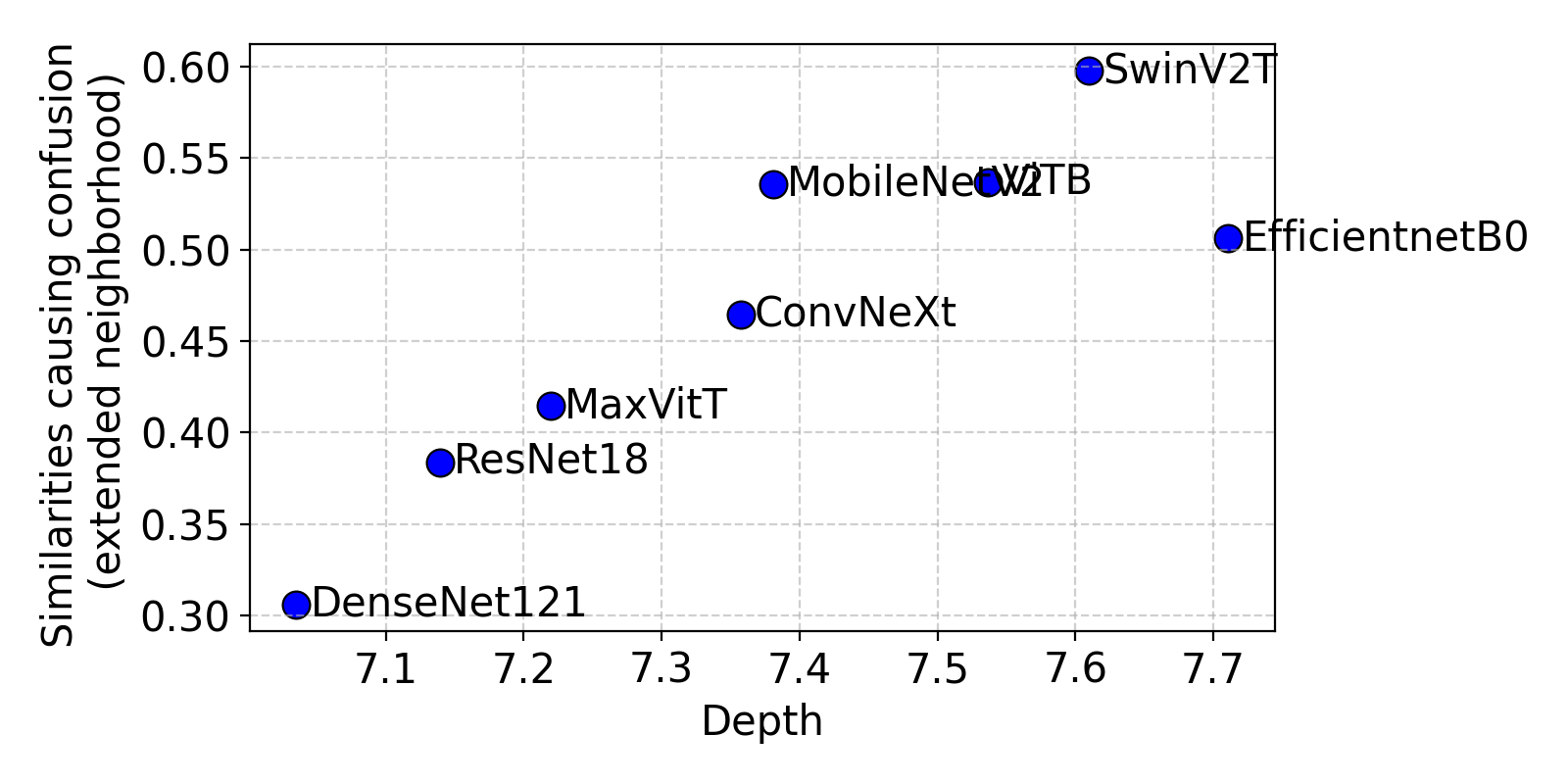}
    }
    \subfloat[299 - extended]{%
        \includegraphics[width=0.45\textwidth]{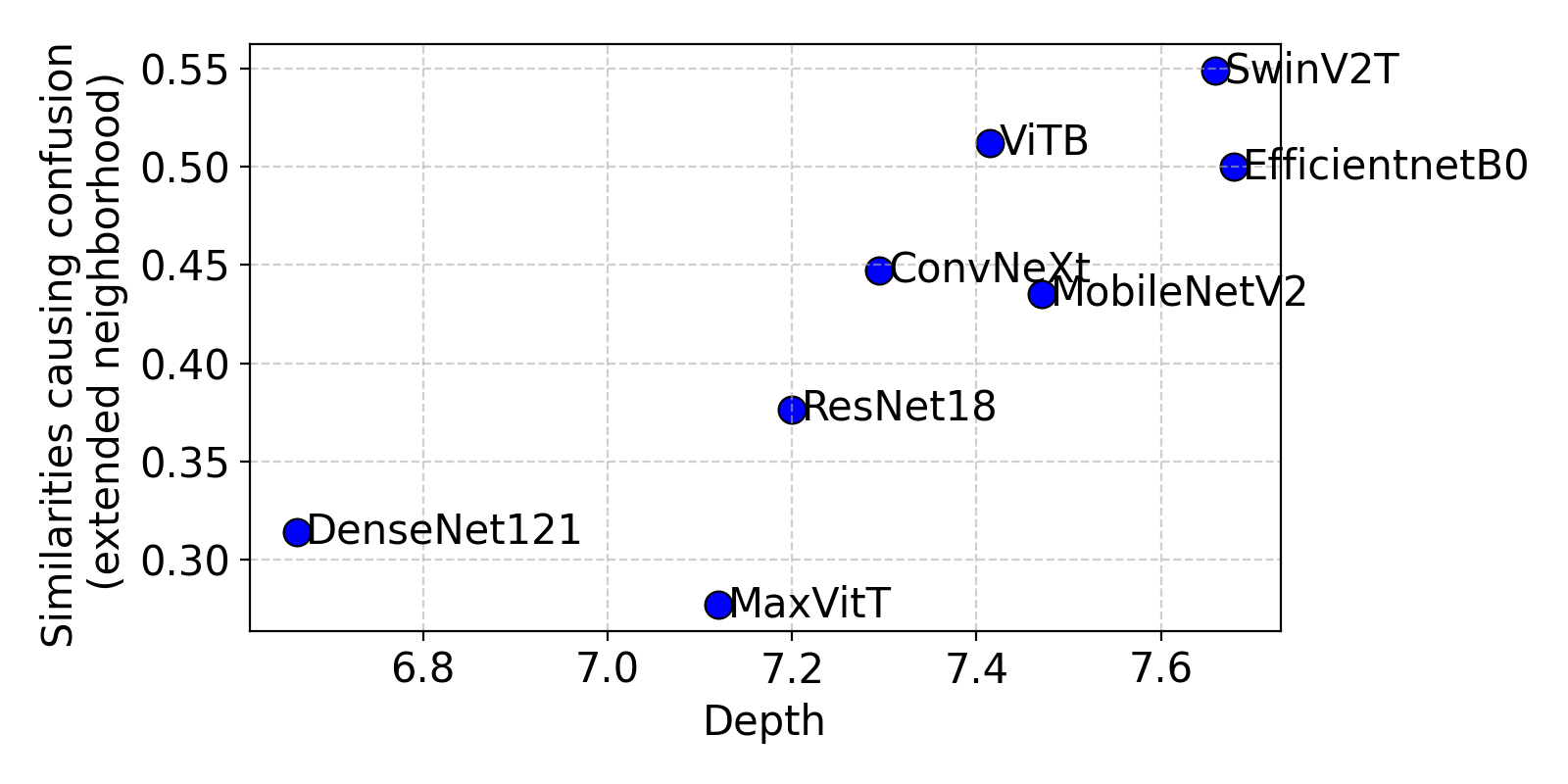}
    }
    \caption{ Similarity Graph Compliance (\textbf{SGC}) as a function of \textbf{SD}: source - NSSG (structural), target - NFSG (functional). Perceived similarities causing confusions.}
    \label{fig:mini_similarities}
\end{figure}

\paragraph{Relationship between SD and Similarity Graph Compliance}
Fig.~\ref{fig:mini_confusions} and  Fig.~\ref{fig:mini_similarities} demonstrate a proportional relationship between network semantic depth and the percentage of explained confusions. Networks with deeper semantic perceptions exhibit more mistakes that can be explained by perceived semantic similarities, a relationship notably pronounced in the \textit{Similarities causing confusions} scenario. This highlights that deeper semantic perception within networks often contributes directly to confusion between classes similar from the model perspective (with similarity estimated from model weights).

\begin{figure}[h!]
    \centering
    \subfloat[Subgraph connected mostly to marine life.]{%
        \includegraphics[width=0.65\textwidth]{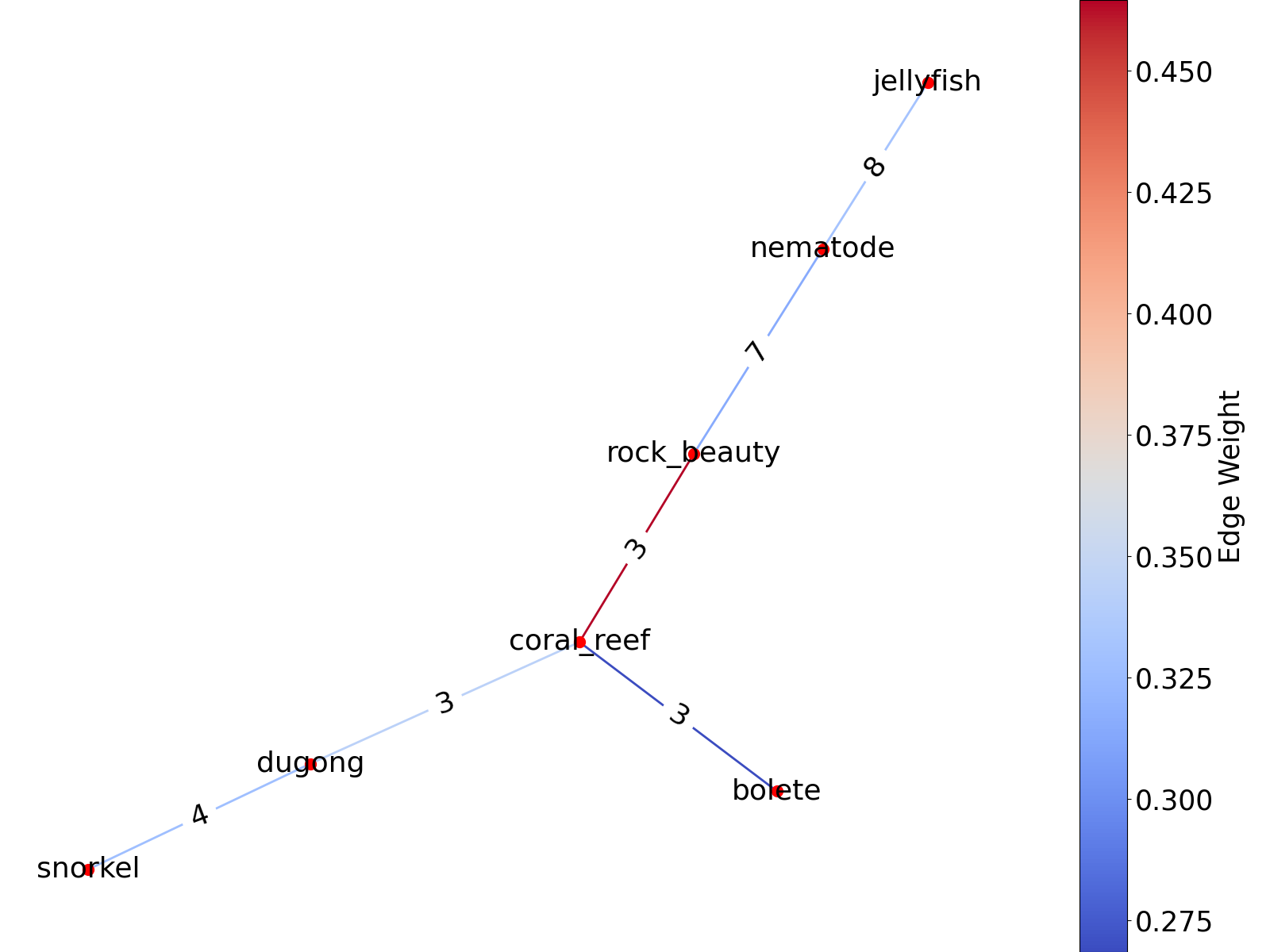}
    }
\\[-2ex]
    \subfloat[S: Coral Reef, T: Rock Beauty]{%
        \includegraphics[width=0.22\textwidth]{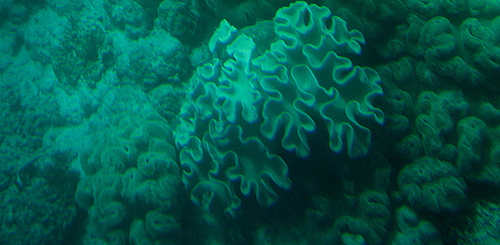}
    }
    \subfloat[S: Rock Beauty, T: Coral Reef]{%
        \includegraphics[width=0.22\textwidth]{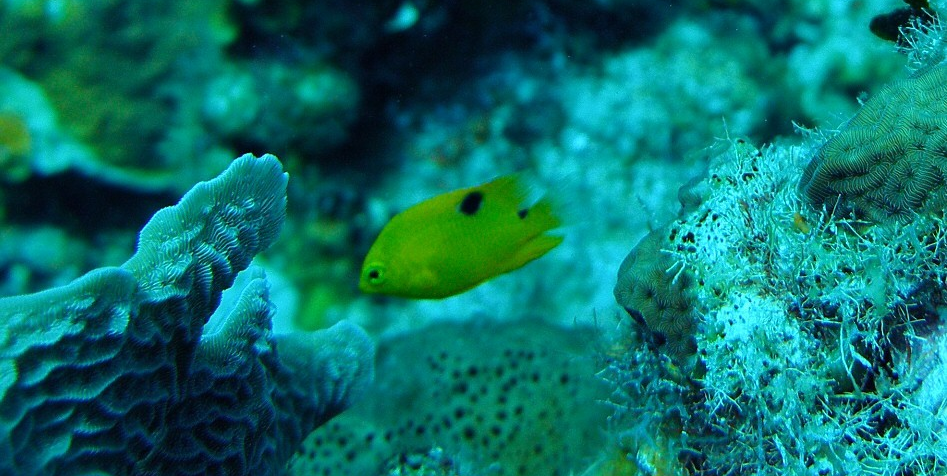}
    }
\quad
    \subfloat[S: Coral Reef, T: Rock Beauty]{%
        \includegraphics[width=0.22\textwidth]{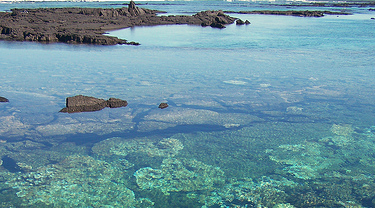}
    }
    \subfloat[S: Rock Beauty, T: Coral Reef]{%
        \includegraphics[width=0.22\textwidth]{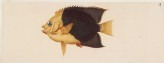}
    }
\\[-2ex]
    \subfloat[S: Coral Reef,\\ T: Bolete]{%
        \includegraphics[width=0.22\textwidth]{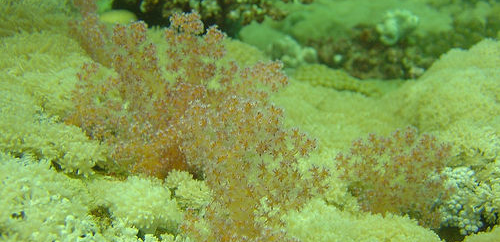}
    }
    \subfloat[S: Bolete,\\ T: Coral Reef]{%
        \includegraphics[width=0.22\textwidth]{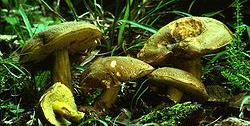}
    }
\quad
    \subfloat[S: Coral Reef,\\ T: Bolete]{%
        \includegraphics[width=0.22\textwidth]{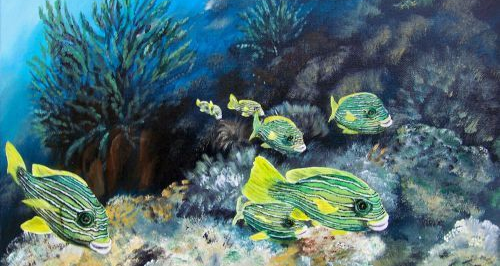}
    }
    \subfloat[S: Bolete,\\ T: Coral Reef]{%
        \includegraphics[width=0.22\textwidth]{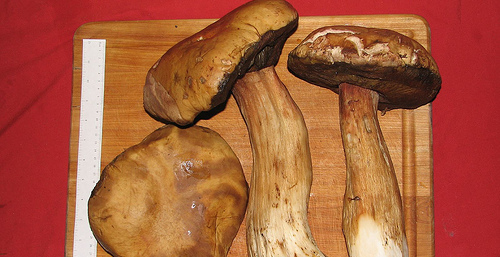}
    }
    \caption{Example subgraph for EfficientNetV2B0 and Affinity Clustering. S - Source, T- Target. Max supporting images on the left, min - on the right. Images taken from the mini-ImageNet dataset.}
    \label{fig:EfficientNetV2B0}
\end{figure}

\paragraph{Visual analysis examples} The example subgraph presented in Fig.~\ref{fig:EfficientNetV2B0} for EfficientNetV2B0 model shows semantic connections predominantly related to marine environments (e.g., coral reef, snorkel, jellyfish). Further analysis of 2 out of 3 pairs with a semantically irrelevant score (SD=3, see Tab. \ref{table:example_classes_mini_image_net}) with the proposed visualization method highlights that the inspected network perceives a coral reef as similar to a rock beauty fish, possibly because they are often captured together. Both categories are also connected via a different semantic relation than similarity, namely containment (coral reefs are a habitat of rock beauties). It is supported by the fact that the least supporting images are a coral reef photo taken from a distance and a fish outside of water. In the case of the bolete-coral reef pair, the resemblance is most likely due to the purely visual similarity of boletes growing in groups to corals (the basic building block of reefs), which also grow in groups. This analysis illustrates the ability of semantic graphs to reveal non-similarity semantic relationships like co-occurrence and containment.

\begin{figure}[h!]
    \centering
    \subfloat[An example subgraph.]{%
        \includegraphics[width=0.65\textwidth]{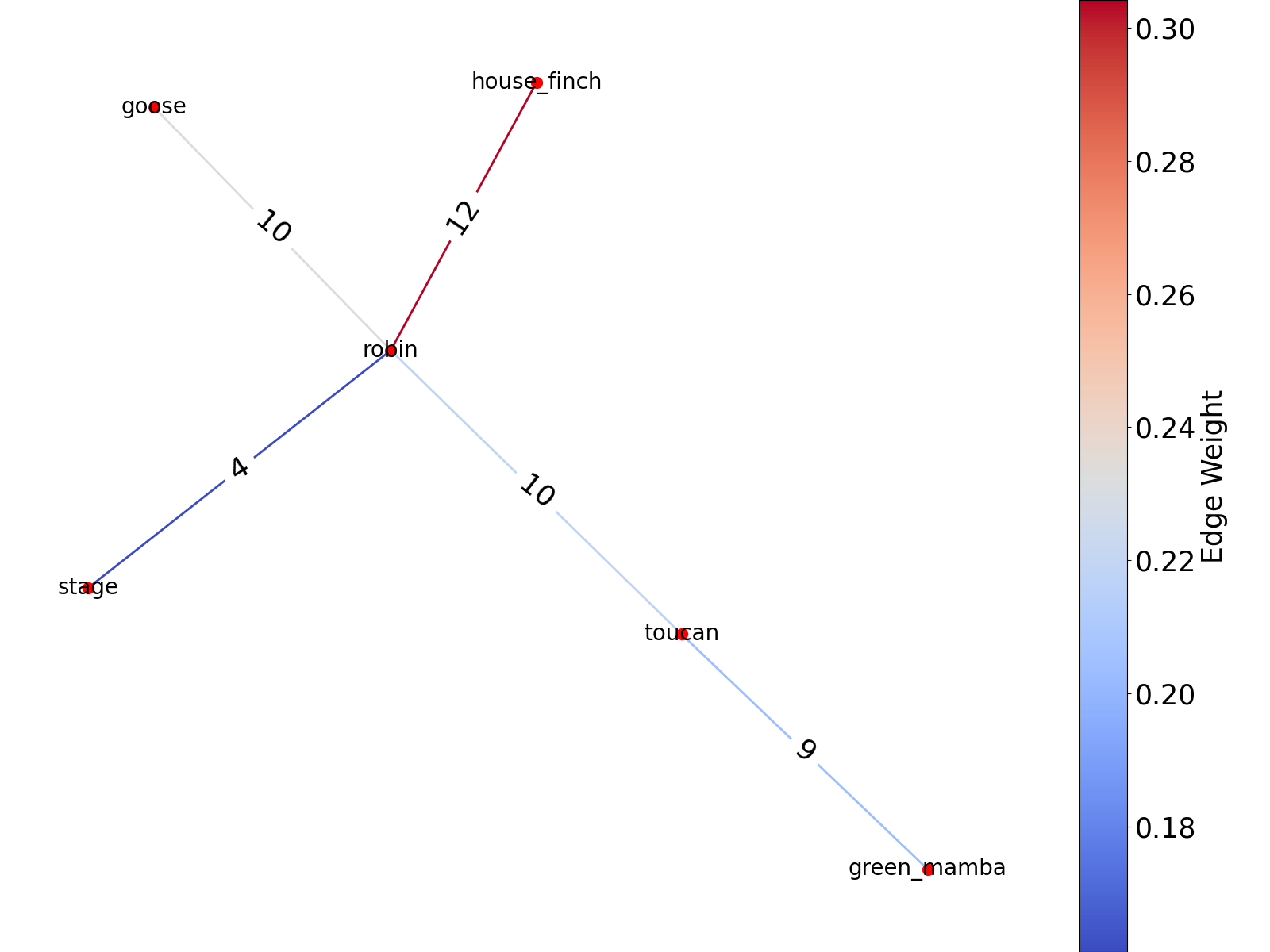}
    }
\\[-2ex]
    \subfloat[S: Green Mamba, T: Toucan]{%
        \includegraphics[width=0.22\textwidth]{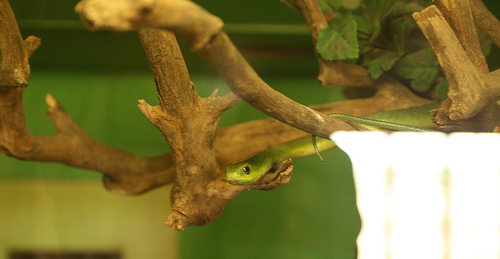}
    }
    \subfloat[S: Toucan, T: Green Mamba]{%
        \includegraphics[width=0.22\textwidth]{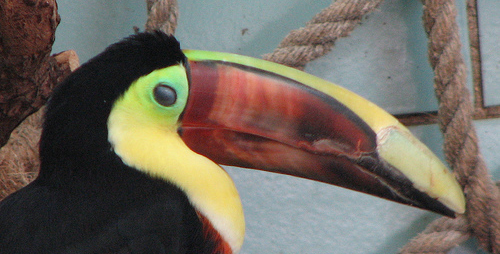}
    }
\quad
    \subfloat[S: Green Mamba, T: Toucan]{%
        \includegraphics[width=0.22\textwidth]{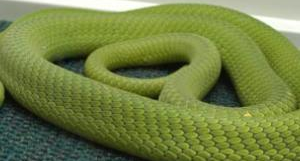}
    }
    \subfloat[S: Toucan, T: Green Mamba]{%
        \includegraphics[width=0.22\textwidth]{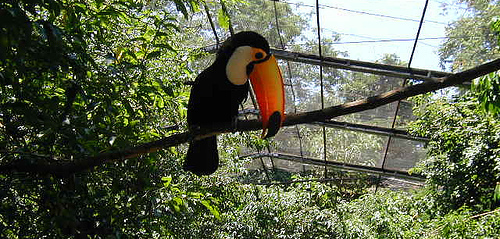}
    }
\\[-2ex]
    \subfloat[S: Robin,\\ T: Stage]{%
        \includegraphics[width=0.22\textwidth]{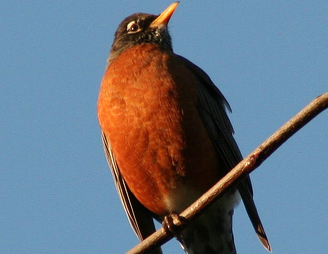}
    }
    \subfloat[S: Stage,\\ T: Robin]{%
        \includegraphics[width=0.22\textwidth]{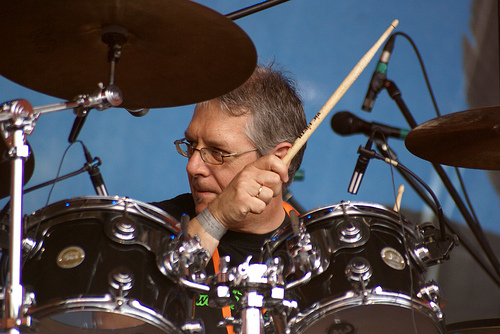}
    }
\quad
    \subfloat[S: Robin,\\ T: Stage]{%
        \includegraphics[width=0.22\textwidth]{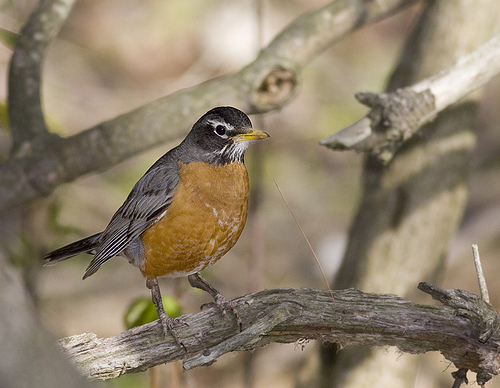}
    }
    \subfloat[S: Stage,\\ T: Robin]{%
        \includegraphics[width=0.22\textwidth]{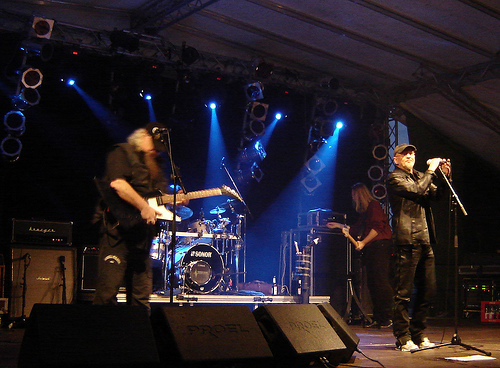}
    }
    \caption{Example subgraph for MaxVitT and Affinity Clustering. S - Source, T- Target. Max supporting images on the left, min - on the right. Images taken from the mini-ImageNet dataset.}
    \label{fig:MaxVitT}
\end{figure}

Fig.~\ref{fig:MaxVitT} presents an example subgraph obtained for the MaxVitT model, predominantly connecting birds, e.g., toucan, and robin. Analysis of pairs with the lowest semantic similarities, toucan-green mamba or robin-stage, suggests that high similarity between categories is caused by the co-occurrence of green markings in both animals as well as the similarity of environments in which they can be found. For the robin-stage case, the similarity can be caused by the common background of their training images (sky) and the similarity of some instruments to branches on which robins are often photographed. Nevertheless, the low depth value of their common ancestor indicates a potential problem with the network similarity structure. This illustrates how our method can be used to inspect models for the purpose of finding deficiencies in the perception of the world caused by purely visual similarities.

\begin{figure}[h!]
    \centering
    \subfloat[An example subgraph.]{%
        \includegraphics[width=0.65\textwidth]{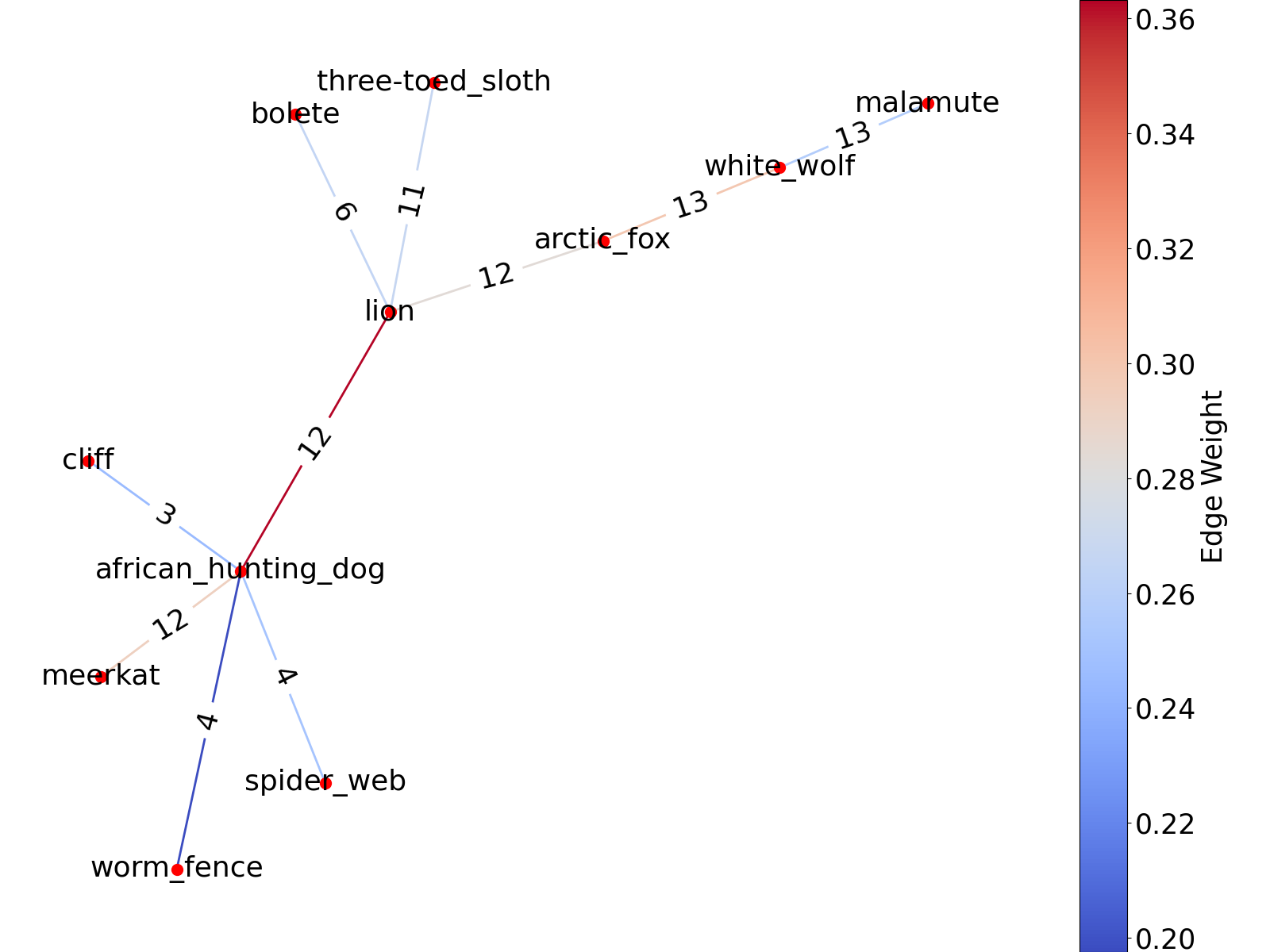}
    }
\\[-2ex]
    \subfloat[S: Lion,\\ T: Arctic Fox]{%
        \includegraphics[width=0.2\textwidth]{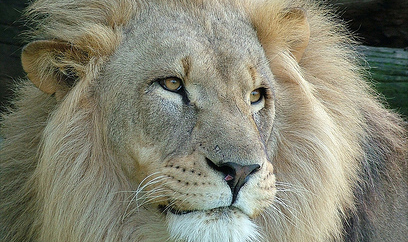}
    }
    \subfloat[S:Arctic Fox,\\ T: Lion]{%
        \includegraphics[width=0.2\textwidth]{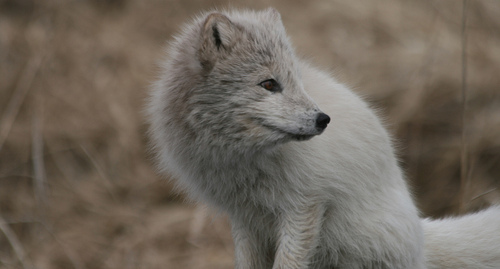}
    }
\quad
    \subfloat[S: Lion,\\ T: Arctic Fox]{%
        \includegraphics[width=0.2\textwidth]{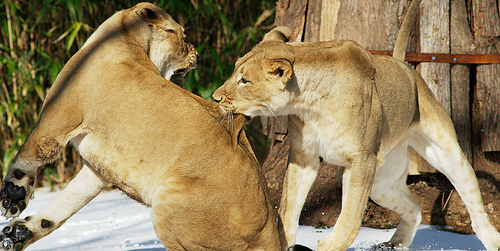}
    }
    \subfloat[S: Arctic Fox,\\ T: Lion]{%
        \includegraphics[width=0.2\textwidth]{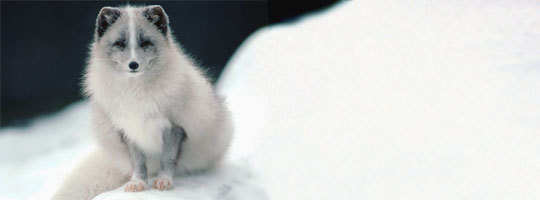}
    }
\\[-2ex]
    \subfloat[S:Arctic Fox, \\T: White Wolf]{%
        \includegraphics[width=0.2\textwidth]{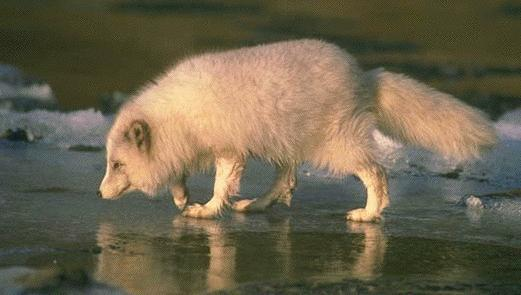}
    }
    \subfloat[S: White Wolf,\\ T: Arctic Fox]{%
        \includegraphics[width=0.2\textwidth]{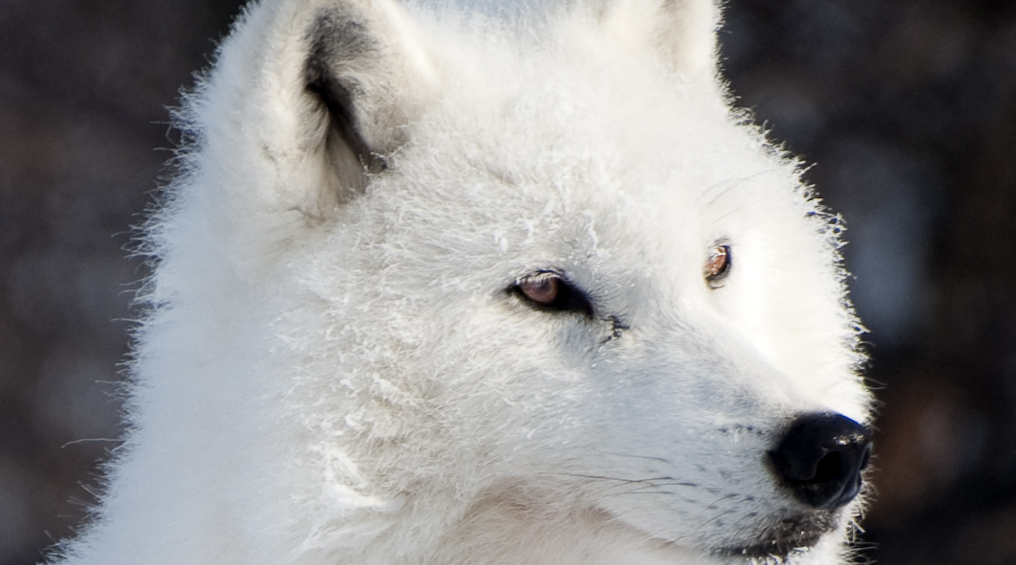}
    }
\quad
    \subfloat[S: Arctic Fox,\\ T: White Wolf]{%
        \includegraphics[width=0.2\textwidth]{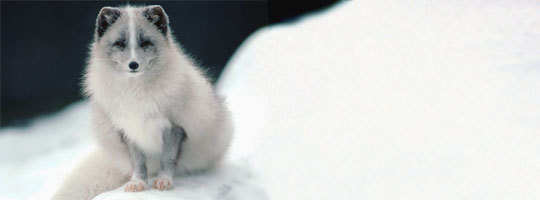}
    }
    \subfloat[S: White Wolf,\\ T: Arctic Fox]{%
        \includegraphics[width=0.2\textwidth]{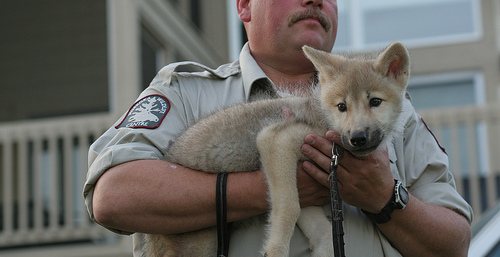}
    }
    \caption{Example subgraph for ConveXtT and Affinity Clustering. S - Source, T- Target. Max supporting images are placed on the left, min - on the right. Images taken from the mini-ImageNet dataset.}
    \label{fig:ConveXtT}
\end{figure}

Fig.~\ref{fig:ConveXtT} presents an example subgraph obtained for ConveXtT model, predominantly connected to different animals. This example shows how the proposed visualization can be used to  locate classes with high semantic similarities and find their instances that support and illustrate this similarity. The example presents how foxes can be similar to lions due to their long, brown/grayish fur in summer, and to white wolves in winter due to their fluffy, white fur during this season. The example illustrates how the arrangement of similar categories can visually demonstrate transitions between classes. For instance, the path from hunting dog to malamute is intuitively interpretable due to the intermediate categories and their visual characteristics (hunting dog $\longrightarrow$ lion $\longrightarrow$ arctic fox $\longrightarrow$ white wolf $\longrightarrow$ malamute). Such transitions can help us hypothesize about about the network’s internal representations and perception of class similarities.

\section{Discussion}
\textbf{SD for different models} While focusing on the analysis of supervised models, we applied our method to models trained with different paradigms, such as self-supervised learning or text-image tasks, DINOv2, and CLIP, respectively. As without modification, these models do not perform classification, to generate the Network Structural Similarity Graph (NSSG), either (1) they had to be extended with a linear classification layer and retrained, (2) aggregated features extracted by the model have to be used to compute all the pair-wise similarities. We have chosen the second approach. Table~\ref{table:mini_clip_dino_sd} presents the SDs obtained for these models and mini-ImageNet. Results indicate that these networks also develop a semantic depth perception at a high level of semantic specificity (app. 7.5-7.9), which is even deeper than for the majority of supervised models. While such behavior could be expected of CLIP, which also uses text sources for training (with intrinsic lexical relations), the high depth of semantic perception is surprising for DINOv2, which is not constrained to learn explicit semantic categories during training either by classes of supervised training or by textual-visual connections of text-image models. 

\textbf{Practical implications} Our results show the proposed framework enables numerical and visual analysis of networks' semantic structures and error predictability. The proposed method is data-free and has high performance: computing SD for a 100x100 matrix takes less than 100ms for all the clustering methods. This makes it viable to compute SD even during training. Examining SD during model training reveals rapid initial growth, with subsequent declines likely due to loss optimization, where the model abandons the more natural perception of similarities, adapting to the training examples. This opens up the predictive potential of SD as a training interpretability indicator. 
 
\begin{table}
\caption{Example mini-ImageNet classes with their WordNet paths to the root (entity). Nodes at level 7-9 represent specific categories - it is a semantically relevant level with rich information content.}
\label{table:example_classes_mini_image_net}
\centering
\scriptsize
\begin{tabular}{l|l|l|l}
\textbf{Depth} & \textbf{golden retriever} & \textbf{garbage truck} & \textbf{coral reef} \\ \hline
1 & entity & entity & entity \\ \hline
2 & physical entity & physical entity & physical entity \\ \hline
3 & object & object & object \\ \hline
4 & whole & whole & geological formation \\ \hline
5 & living thing & artifact & natural elevation  \\ \hline
6 & organism & instrumentality & ridge  \\ \hline
7 & animal & container & reef  \\ \hline
8 & chordate & wheeled vehicle & \textbf{coral reef}  \\ \hline
9 & vertebrate & self-propelled vehicle &  \\ \hline
10 & mammal & motor vehicle &  \\ \hline
11 & placental & truck &  \\ \hline
12 & carnivore & \textbf{garbage truck} &  \\ \hline
13 & canine &  &  \\ \hline
14 & dog &  &  \\ \hline
15 & hunting dog &  &  \\ \hline
16 & sporting dog &  &  \\ \hline
17 & retriever &  &  \\ \hline
18 & \textbf{golden retriever} &  &  \\ 
\end{tabular}
\end{table}

\begin{table}[h]
\caption{Semantic Depth (SD) values obtained for Self-Supervised Learning and text-image models.}
\label{table:mini_clip_dino_sd}
\centering
\begin{tabular}{l|l|l|l|l|l}
Network & Louvain & OPTICS & Affinity & HDBSCAN & No clustering \\
\hline
DINOv2 & 7.75 (4) & 7.55 (7) & 7.60 (16) & 7.68 (3) & 7.68 (1) \\
CLIP & 7.79 (4) & 7.58 (3) & 7.87 (14) & 7.62 (4) & 7.82 (1) \\
\end{tabular}
\end{table}

\section{Conclusions}

Our findings demonstrate that established deep vision neural networks inherently encode highly specific semantic hierarchies, confirming that their perception extends beyond general concepts to deeper, detailed categorizations. The proposed framework, which leverages semantic depth (SD) and similarity graph compliance metrics anchored in WordNet hierarchies, effectively reveals nuanced model perceptions and error predictability without requiring additional data, relying solely on network weights. Experiments illustrate that higher semantic depths correlate with increased predictability of network mistakes, especially pronounced in higher-resolution datasets like Mini-ImageNet. Our visualization tools further enhance interpretability, allowing intuitive identification of semantic relationships and misperceptions. Consequently, this framework offers significant practical value for debugging visual classifiers, enhancing transparency, and guiding model refinement.

\section*{Acknowledgements}
We gratefully acknowledge Polish high-performance computing infrastructure PLGrid (HPC Center: ACK Cyfronet AGH) for providing computer facilities and support within computational grant no. PLG/2024/017513
\bibliographystyle{splncs04}
\bibliography{semantic_depth_matters}

@article{bib:rosch1978cognition,
  title={Cognition and categorization.},
  author={Rosch, Eleanor Ed and Lloyd, Barbara B},
  year={1978},
  publisher={Lawrence Erlbaum}
}

@inproceedings{bib:you2020graph,
  title={Graph structure of neural networks},
  author={You, Jiaxuan and Leskovec, Jure and He, Kaiming and Xie, Saining},
  booktitle={International Conference on Machine Learning},
  pages={10881--10891},
  year={2020},
  organization={PMLR}
}

@article{bib:lawson2012population,
  title={Population identification using genetic data},
  author={Lawson, Daniel John and Falush, Daniel},
  journal={Annual review of genomics and human genetics},
  volume={13},
  number={1},
  pages={337--361},
  year={2012},
  publisher={Annual Reviews}
}

@article{bib:kriegeskorte2008rdm,
  title={Representational similarity analysis - connecting the branches of systems neuroscience},
  author={Kriegeskorte, Nikolaus and Mur, Marieke and Bandettini, Peter},
  journal={Frontiers in Systems Neuroscience},
  volume={2},
  number={4},
  year={2008}
}

@inproceedings{bib:williamsequivalence,
  title={Equivalence between representational similarity analysis, centered kernel alignment, and canonical correlations analysis},
  author={Williams, Alex H},
  booktitle={UniReps: 2nd Edition of the Workshop on Unifying Representations in Neural Models}
}

@article{bib:oquab2023dinov2,
  title={Dinov2: Learning robust visual features without supervision},
  author={Oquab, Maxime and Darcet, Timoth{\'e}e and Moutakanni, Th{\'e}o and Vo, Huy and Szafraniec, Marc and Khalidov, Vasil and Fernandez, Pierre and Haziza, Daniel and Massa, Francisco and El-Nouby, Alaaeldin and others},
  journal={arXiv preprint arXiv:2304.07193},
  year={2023}
}

@inproceedings{bib:radford2021learning,
  title={Learning transferable visual models from natural language supervision},
  author={Radford, Alec and Kim, Jong Wook and Hallacy, Chris and Ramesh, Aditya and Goh, Gabriel and Agarwal, Sandhini and Sastry, Girish and Askell, Amanda and Mishkin, Pamela and Clark, Jack and others},
  booktitle={International conference on machine learning},
  pages={8748--8763},
  year={2021},
  organization={PMLR}
}

@inproceedings{bib:purvine2023experimental,
  title={Experimental observations of the topology of convolutional neural network activations},
  author={Purvine, Emilie and Brown, Davis and Jefferson, Brett and Joslyn, Cliff and Praggastis, Brenda and Rathore, Archit and Shapiro, Madelyn and Wang, Bei and Zhou, Youjia},
  booktitle={AAAI Conference on Artificial Intelligence},
  volume={37},
  number={8},
  pages={9470--9479},
  year={2023}
}

@article{bib:selvaraju2016grad,
  title={Grad-CAM: Why did you say that?},
  author={Selvaraju, Ramprasaath R and Das, Abhishek and Vedantam, Ramakrishna and Cogswell, Michael and Parikh, Devi and Batra, Dhruv},
  journal={arXiv preprint arXiv:1611.07450},
  year={2016}
}

@inproceedings{bib:zintgraf2017visualizing,
  title={Visualizing Deep Neural Network Decisions: Prediction Difference Analysis},
  author={Zintgraf, Luisa M and Cohen, Taco S and Adel, Tameem and Welling, Max},
  booktitle={International Conference on Learning Representations},
  year={2017}
}

@article{bib:goldstone1994similarity,
  title={Similarity, interactive activation, and mapping.},
  author={Goldstone, Robert L},
  journal={Journal of experimental psychology: learning, memory, and cognition},
  volume={20},
  number={1},
  pages={3},
  year={1994},
  publisher={American Psychological Association}
}

@inproceedings{bib:bertinetto2020making,
  title={Making better mistakes: Leveraging class hierarchies with deep networks},
  author={Bertinetto, Luca and Mueller, Romain and Tertikas, Konstantinos and Samangooei, Sina and Lord, Nicholas A},
  booktitle={{IEEE/CVF Conference on Computer Vision and Pattern Recognition}},
  pages={12506--12515},
  year={2020}
}

@article{bib:bilal2017convolutional,
  title={Do convolutional neural networks learn class hierarchy?},
  author={Bilal, Alsallakh and Jourabloo, Amin and Ye, Mao and Liu, Xiaoming and Ren, Liu},
  journal={{IEEE Transactions on Visualization and Computer Graphics}},
  volume={24},
  number={1},
  pages={152--162},
  year={2017}
}

@inproceedings{bib:yuan2021explaining,
  title={Explaining information flow inside vision transformers using markov chain},
  author={Yuan, Tingyi and Li, Xuhong and Xiong, Haoyi and Cao, Hui and Dou, Dejing},
  booktitle={eXplainable AI approaches for debugging and diagnosis.},
  year={2021}
}

@article{bib:russakovsky2015imagenet,
  title={Imagenet large scale visual recognition challenge},
  author={Russakovsky, Olga and Deng, Jia and Su, Hao and Krause, Jonathan and Satheesh, Sanjeev and Ma, Sean and Huang, Zhiheng and Karpathy, Andrej and Khosla, Aditya and Bernstein, Michael and others},
  journal={{International Journal of Computer Vision}},
  volume={115},
  pages={211--252},
  year={2015}
}

@inproceedings{bib:deng2010does,
  title={What does classifying more than 10,000 image categories tell us?},
  author={Deng, Jia and Berg, Alexander C and Li, Kai and Fei-Fei, Li},
  booktitle={European Conference on Computer Vision},
  pages={71--84},
  year={2010}
}

@article{bib:huang2021semantic,
  title={Semantic relatedness emerges in deep convolutional neural networks designed for object recognition},
  author={Huang, Taicheng and Zhen, Zonglei and Liu, Jia},
  journal={{Frontiers in Computational Neuroscience}},
  volume={15},
  pages={625804},
  year={2021}
}

@inproceedings{bib:mopuri2020adversarial,
  title={Adversarial Fooling Beyond" Flipping the Label"},
  author={Mopuri, Konda Reddy and Shaj, Vaisakh and Babu, R Venkatesh},
  booktitle={{IEEE/CVF Conference on Computer Vision and Pattern Recognition Workshops}},
  pages={778--779},
  year={2020}
}

@inproceedings{bib:filus2023netsat,
  title={NetSat: Network Saturation Adversarial Attack},
  author={Filus, Katarzyna and Domanska, Joanna},
  booktitle={{IEEE International Conference on Big Data}},
  pages={5038--5047},
  year={2023}
}

@article{bib:muttenthaler2024improving,
  title={Improving neural network representations using human similarity judgments},
  author={Muttenthaler, Lukas and Linhardt, Lorenz and Dippel, Jonas and Vandermeulen, Robert A and Hermann, Katherine and Lampinen, Andrew and Kornblith, Simon},
  journal={{Advances in Neural Information Processing Systems}},
  volume={36},
  year={2023}
}

@inproceedings{bib:pedersen2004wordnet,
  title={WordNet:: Similarity-Measuring the Relatedness of Concepts.},
  author={Pedersen, Ted and Patwardhan, Siddharth and Michelizzi, Jason and others},
  year={2004}
}

@article{bib:leacock1998combining,
  title={Combining local context and WordNet similarity for word sense identification},
  author={Leacock, Claudia and Chodorow, Martin},
  journal={{WordNet: An electronic lexical database}},
  volume={49},
  number={2},
  pages={265--283},
  year={1998}
}

@inproceedings{bib:kornblith2019similarity,
  title={Similarity of neural network representations revisited},
  author={Kornblith, Simon and Norouzi, Mohammad and Lee, Honglak and Hinton, Geoffrey},
  booktitle={International conference on machine learning},
  pages={3519--3529},
  year={2019},
  organization={PMLR}
}

@article{bib:filus2024similarity,
  title={Similarity-driven adversarial testing of neural networks},
  author={Filus, Katarzyna and Doma{\'n}ska, Joanna},
  journal={Knowledge-Based Systems},
  volume={305},
  pages={112621},
  year={2024},
  publisher={Elsevier}
}

@book{bird2009natural,
  title={Natural language processing with Python: analyzing text with the natural language toolkit},
  author={Bird, Steven and Klein, Ewan and Loper, Edward},
  year={2009},
  publisher={" O'Reilly Media, Inc."}
}

@article{bib:miller1990introduction,
  title={Introduction to WordNet: An on-line lexical database},
  author={Miller, George A and Beckwith, Richard and Fellbaum, Christiane and Gross, Derek and Miller, Katherine J},
  journal={International journal of lexicography},
  volume={3},
  number={4},
  pages={235--244},
  year={1990},
  publisher={Oxford University Press}
}

@inproceedings{bib:wang2011refining,
  title={Refining the notions of depth and density in wordnet-based semantic similarity measures},
  author={Wang, Tong and Hirst, Graeme},
  booktitle={2011 Conference on Empirical Methods in Natural Language Processing},
  pages={1003--1011},
  year={2011}
}

@inproceedings{bib:wu:1994verb,
    title = "Verb Semantics and Lexical Selection",
    author = "Wu, Zhibiao  and
      Palmer, Martha",
    booktitle = "32nd Annual Meeting of the Association for Computational Linguistics",
    month = jun,
    year = "1994",
    address = "Las Cruces, New Mexico, USA",
    publisher = "Association for Computational Linguistics",
    pages = "133--138"
}

@inproceedings{maxvit_model,
  title={Maxvit: Multi-axis vision transformer},
  author={Tu, Zhengzhong and Talebi, Hossein and Zhang, Han and Yang, Feng and Milanfar, Peyman and Bovik, Alan and Li, Yinxiao},
  booktitle={European conference on computer vision},
  pages={459--479},
  year={2022},
  organization={Springer}
}

@misc{convnext_model,
      title={A ConvNet for the 2020s}, 
      author={Zhuang Liu and Hanzi Mao and Chao-Yuan Wu and Christoph Feichtenhofer and Trevor Darrell and Saining Xie},
      year={2022},
      eprint={2201.03545},
      archivePrefix={arXiv},
      primaryClass={cs.CV},
      url={https://arxiv.org/abs/2201.03545}, 
}

@misc{vit_model,
      title={An Image is Worth 16x16 Words: Transformers for Image Recognition at Scale}, 
      author={Alexey Dosovitskiy and Lucas Beyer and Alexander Kolesnikov and Dirk Weissenborn and Xiaohua Zhai and Thomas Unterthiner and Mostafa Dehghani and Matthias Minderer and Georg Heigold and Sylvain Gelly and Jakob Uszkoreit and Neil Houlsby},
      year={2021},
      eprint={2010.11929},
      archivePrefix={arXiv},
      primaryClass={cs.CV},
      url={https://arxiv.org/abs/2010.11929}, 
}

@misc{swin_model,
      title={Swin Transformer: Hierarchical Vision Transformer using Shifted Windows}, 
      author={Ze Liu and Yutong Lin and Yue Cao and Han Hu and Yixuan Wei and Zheng Zhang and Stephen Lin and Baining Guo},
      year={2021},
      eprint={2103.14030},
      archivePrefix={arXiv},
      primaryClass={cs.CV},
      url={https://arxiv.org/abs/2103.14030}, 
}

@misc{efficientnet_model,
      title={EfficientNet: Rethinking Model Scaling for Convolutional Neural Networks}, 
      author={Mingxing Tan and Quoc V. Le},
      year={2020},
      eprint={1905.11946},
      archivePrefix={arXiv},
      primaryClass={cs.LG},
      url={https://arxiv.org/abs/1905.11946}, 
}

@misc{densenet_model,
      title={Densely Connected Convolutional Networks}, 
      author={Gao Huang and Zhuang Liu and Laurens van der Maaten and Kilian Q. Weinberger},
      year={2018},
      eprint={1608.06993},
      archivePrefix={arXiv},
      primaryClass={cs.CV},
      url={https://arxiv.org/abs/1608.06993}, 
}

@misc{resnet_model,
      title={Deep Residual Learning for Image Recognition}, 
      author={Kaiming He and Xiangyu Zhang and Shaoqing Ren and Jian Sun},
      year={2015},
      eprint={1512.03385},
      archivePrefix={arXiv},
      primaryClass={cs.CV},
      url={https://arxiv.org/abs/1512.03385}, 
}

@misc{mobilenet_model,
      title={MobileNetV2: Inverted Residuals and Linear Bottlenecks}, 
      author={Mark Sandler and Andrew Howard and Menglong Zhu and Andrey Zhmoginov and Liang-Chieh Chen},
      year={2019},
      eprint={1801.04381},
      archivePrefix={arXiv},
      primaryClass={cs.CV},
      url={https://arxiv.org/abs/1801.04381}, 
}

@inproceedings{mini_imagenet,
 author = {Vinyals, Oriol and Blundell, Charles and Lillicrap, Timothy and kavukcuoglu, koray and Wierstra, Daan},
 booktitle = {Advances in Neural Information Processing Systems},
 editor = {D. Lee and M. Sugiyama and U. Luxburg and I. Guyon and R. Garnett},
 pages = {},
 publisher = {Curran Associates, Inc.},
 title = {Matching Networks for One Shot Learning},
 url = {https://proceedings.neurips.cc/paper_files/paper/2016/file/90e1357833654983612fb05e3ec9148c-Paper.pdf},
 volume = {29},
 year = {2016}
}

@article{cifar,
author = {Krizhevsky, Alex},
year = {2012},
month = {05},
pages = {},
title = {Learning Multiple Layers of Features from Tiny Images},
journal = {University of Toronto}
}

@misc{adamW,
      title={Decoupled Weight Decay Regularization}, 
      author={Ilya Loshchilov and Frank Hutter},
      year={2019},
      eprint={1711.05101},
      archivePrefix={arXiv},
      primaryClass={cs.LG},
      url={https://arxiv.org/abs/1711.05101}, 
}

@misc{rop_scheduler,
      title={A Simple Dynamic Learning Rate Tuning Algorithm For Automated Training of DNNs}, 
      author={Koyel Mukherjee and Alind Khare and Ashish Verma},
      year={2019},
      eprint={1910.11605},
      archivePrefix={arXiv},
      primaryClass={cs.LG},
      url={https://arxiv.org/abs/1910.11605}, 
}

\end{document}